\documentclass[]{oppo}
\usepackage[toc,page,header]{appendix}

\usepackage{latexsym}
\usepackage[T1]{fontenc}
\usepackage[utf8]{inputenc}
\usepackage{microtype}
\usepackage{inconsolata}
\usepackage{graphicx}
\usepackage{hyperref}       % hyperlinks
\usepackage{url}            % simple URL typesetting
\usepackage{booktabs}       % professional-quality tables
\usepackage{amsfonts}       % blackboard math symbols
\usepackage{nicefrac}       % compact symbols for 1/2, etc.
\usepackage{stackengine}
\usepackage{microtype}      % microtypography
\usepackage{colortbl}
\usepackage{xcolor}
\usepackage{amsmath}
\usepackage{amssymb}
\usepackage{amsthm}
\usepackage{mathrsfs}
\usepackage{pifont}
\usepackage{MnSymbol}
\usepackage{balance}
\usepackage{enumitem}
\usepackage{listings}
\usepackage{xcolor}
\usepackage{natbib}
\usepackage{multicol}
\usepackage{xspace}

\AtBeginDocument{%
  \providecommand\BibTeX{{%
    \normalfont B\kern-0.5em{\scshape i\kern-0.25em b}\kern-0.8em\TeX}}}

\makeatletter
\DeclareRobustCommand\onedot{\futurelet\@let@token\@onedot}
\def\@onedot{\ifx\@let@token.\else.\null\fi}

\usepackage{setspace}
\usepackage{mathtools}

\usepackage{multirow,booktabs}
\usepackage{subcaption}

\newcommand{\owo}[1]{\textsc{OAgents}}

\definecolor{lightgreen}{RGB}{144, 238, 144} 
\definecolor{lightred}{RGB}{255, 105, 97}

\usepackage{amsmath}
\usepackage{amssymb}
\usepackage{mathtools}
\usepackage{amsthm}

\usepackage{latexsym}
\usepackage[T1]{fontenc}
\usepackage[utf8]{inputenc}
\usepackage{microtype}
\usepackage{inconsolata}
\usepackage{graphicx}
\usepackage{multirow}
\usepackage{booktabs}
\usepackage{pifont}
\usepackage{booktabs}  
\usepackage{tabularx}  
\usepackage{ragged2e}  

\usepackage{tcolorbox}

\definecolor{ogreen}{RGB}{34, 139, 34}

\newtcolorbox{promptbox}[2][Prompt]{
colback=black!5!white,
arc=5pt, 
boxrule=0.5pt,
fonttitle=\bfseries,
title=#1, 
before upper={\small}, fontupper=\fontfamily{ptm}\selectfont,
colframe=#2, 
}
\newcommand{\tcode}[1]{\texttt{\small #1}}

\renewcommand{\arraystretch}{1.3}
\newcommand{\themodel}{\textbf{EcoGym}\xspace}
\usepackage{wrapfig}
\usepackage{adjustbox}
\usepackage{caption}

\theoremstyle{plain}

\theoremstyle{definition}

\theoremstyle{remark}

\usepackage[textsize=tiny]{todonotes}
\usepackage[textsize=tiny]{todonotes}
\usepackage{fontawesome}

\title{EcoGym: Evaluating LLMs for Long-Horizon \\ Plan-and-Execute in Interactive Economies}

\affiliation{OPPO AI Agent Team}

\date{\today}
\correspondence{He Zhu at \email{zhuhe@oppo.com}, Wangchunshu Zhou at \email{zhouwangchunshu@oppo.com}}
\checkdata[Code]{\url{https://github.com/OPPO-PersonalAI/EcoGym}}

\begin{document}
\abstract{
Long-horizon planning is widely recognized as a core capability of autonomous LLM-based agents; however, current evaluation frameworks suffer from being largely episodic, domain-specific, or insufficiently grounded in persistent economic dynamics. We introduce \themodel, a generalizable benchmark for continuous plan-and-execute decision making in interactive economies. \themodel comprises three diverse environments: \textit{Vending} (adapted from the closed-source Vending-Bench, with full \textbf{open-source} release), \textit{Freelance} (\textbf{new}), and \textit{Operation} (\textbf{new}), implemented in a unified decision-making process with standardized interfaces, and budgeted actions over an effectively unbounded horizon (1000+ steps if 365 day-loops for evaluation). The evaluation of \themodel is based on business-relevant outcomes (e.g., net worth, income, and DAU), targeting long-term strategic coherence and robustness under partial observability and stochasticity. Experiments across eleven leading LLMs expose a systematic tension: no single model dominates across all three scenarios. Critically, we find that models exhibit significant suboptimality in either high-level strategies or efficient actions executions. \themodel is released as an open, extensible testbed for transparent long-horizon agent evaluation and for studying controllability–utility trade-offs in economic settings.

}

\maketitle

\section{Introduction}
Long-horizon planning has been recognized as a central agentic capability since the emergence of large language model (LLM)-based agents~\citep{zhang2025landscapeagenticreinforcementlearning,luo2025largelanguagemodelagent}, motivating both agent scaffolds~\citep{wang2025openhandssoftwareagentsdk,li2025agentoriented} and foundation models~\citep{tongyideepresearchteam2025tongyideepresearchtechnicalreport,hu2025stepdeepresearchtechnicalreport} to prioritize long-term navigation and decision-making. This emphasis is likewise reflected in contemporary benchmarks and evaluation protocols, which across a broad range of domains, ranging from deep research~\citep{chen2025xbenchtrackingagentsproductivity,wei2025browsecompsimplechallengingbenchmark} and embodied intelligence~\citep{wang-etal-2022-scienceworld,shridhar2021alfworldaligningtextembodied} to autonomous driving~\citep{jiang2025alphadriveunleashingpowervlms,chen2024pcabenchevaluatingmultimodallarge} and strategic games~\citep{ma2024agentboardanalyticalevaluationboard}, \textit{explicitly} or \textit{implicitly} assess an agent’s capacity for sustained, long-range planning.

\begin{figure}[t]
    \centering

    \begin{subfigure}{0.32\linewidth}
        \centering
        \includegraphics[width=\linewidth]{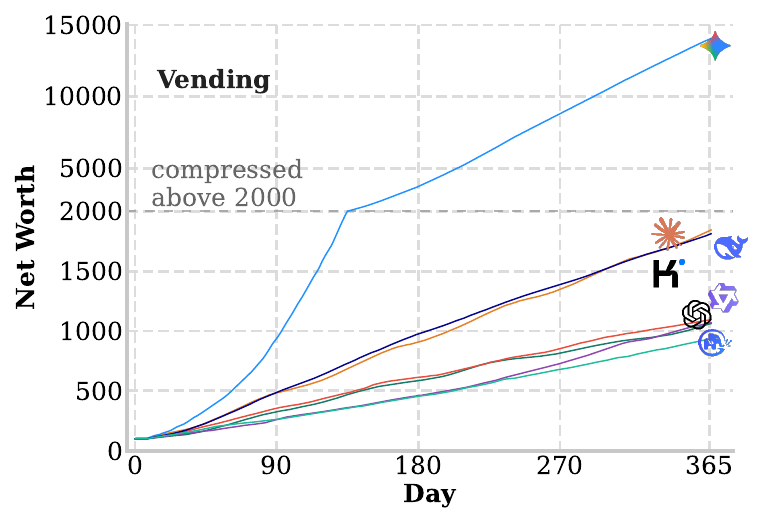}
    \end{subfigure}
    \begin{subfigure}{0.32\linewidth}
        \centering
        \includegraphics[width=\linewidth]{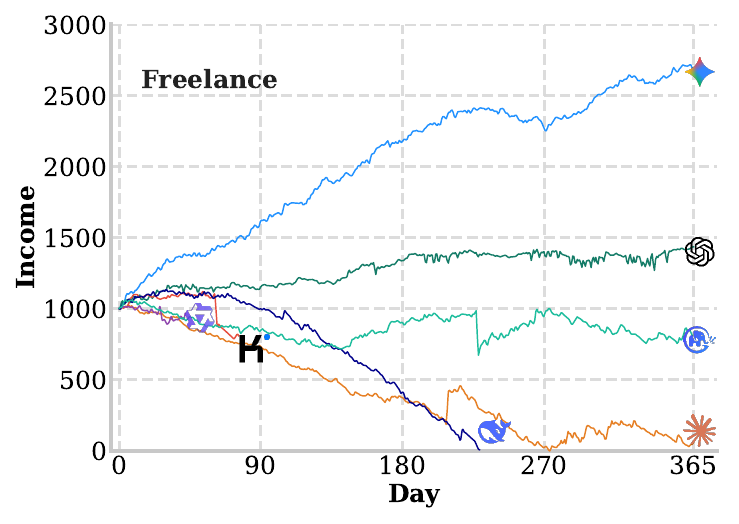}
    \end{subfigure}
    \begin{subfigure}{0.32\linewidth}
        \centering
        \includegraphics[width=\linewidth]{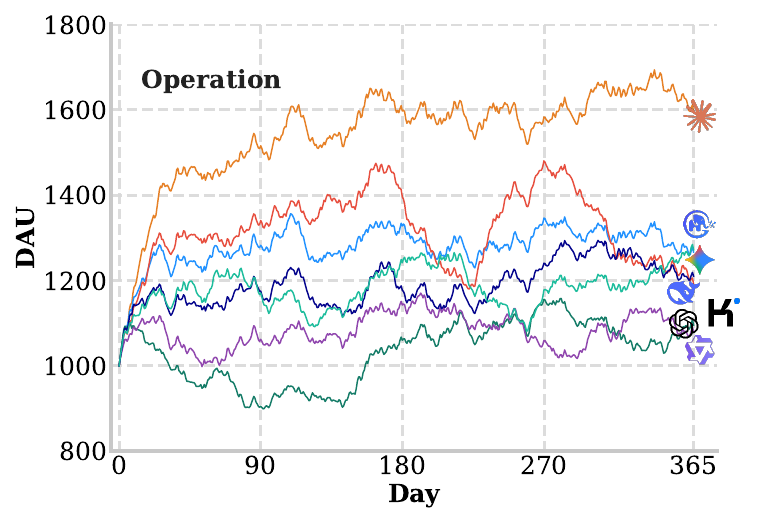}
    \end{subfigure}
    \begin{subfigure}{\linewidth}
        \centering
        \includegraphics[width=\linewidth]{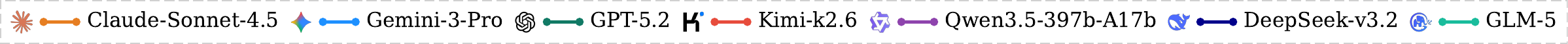}
    \end{subfigure}
    \caption{\textbf{Long-horizon performances across three environments in \themodel.} The plots illustrate the daily progression of key metrics: Net Worth in \textit{Vending} (left), Income in \textit{Freelance} (middle), and DAU in \textit{Operation} (right). \textbf{Note:} Truncated lines represent agents that failed to survive the full simulation horizon due to triggering failure conditions. Only top-performance models are kept for clarity; full experimental results are available in Table~\ref{tab:main_results}.}
    \label{fig:intro}
    \vspace{-10pt}
\end{figure}

Despite the diversity of planning evaluation methodologies, recent research has increasingly situated agents within complex commercial environments and assessed planning competence using tangible, economic returns~\citep{patwardhan2025gdpval,chen2025stockbenchllmagentstrade}. This shift is driven by two complementary concerns: (1) the gap between performance in controlled, virtual benchmarks and robustness under continuous, high-stakes practical deployment, where agents often exhibit fragility despite strong benchmark results; and (2) the greater value of evaluations grounded in economic impact rather than simple rewards. Representative efforts include OpenAI’s GDP Eval~\citep{patwardhan2025gdpval}, Scale AI’s Remote Labor Index~\citep{mazeika2025remote}, and the 32-hour expert-level tasks introduced in RE-Bench~\citep{wijk2024re}, which collectively aim to quantify the macroeconomic potential of such agents.

Notwithstanding recent efforts to benchmark long-horizon planning in commercial settings, ranging from macroeconomic evaluations such as GDPval~\citep{patwardhan2025gdpval} to microeconomic and gamified environments including Vending Bench v1\&v2~\citep{backlund2025vending,vendingbench2} and HeroBench~\citep{anokhin2025herobench}, which introduce competitive pricing or RPG-style resource management, existing testbeds remain limited. They are largely confined to narrow settings such as vending or stylized game scenarios, and therefore fail to reflect the heterogeneous, interdependent business processes characteristic of open-ended economic activity. Moreover, these environments often rely on closed, cumbersome evaluation pipelines (\textit{e.g.}, Vending Bench adheres to a strictly proprietary assessment protocol), underscoring the need for a transparent, community-driven benchmark framework.

To address this, we present \themodel, an interactive environment designed to evaluate LLMs on long-horizon plan-and-execute tasks within interactive economies. Building upon the foundational methodology of Vending Bench, \themodel expends the specific vending scenario into three widely-existed economic settings with a unified interface: \textit{Vending} (adapted from the closed-source Vending-Bench, with full \textbf{open-source} release), \textit{Freelance} (\textbf{new}), and \textit{Operation} (\textbf{new}). It is designed with a theoretically infinite horizon (1000+ steps if 365 day-loops for evaluation), latent economic mechanics, simulating a ``business-never-sleeps'' ecosystem where agents must manage resources sustainably and discover hidden mechanics proactively rather than maximizing short-term rewards. 

Our empirical evaluation on \themodel reveals a significant performance gap in current LLMs: no single model consistently achieves superior performance across all scenarios, highlighting the inherent difficulty of long-horizon economic decision-making. Critically, we find that models exhibit significant suboptimality in either high-level strategies or efficient actions executions. Furthermore, we conduct a comprehensive suite of 8 diagnostic experiments or case studies, encompassing factors such as context window length, agent behavior patterns, additional memory modules, and human baselines. These analyses provide a holistic perspective on the limitations and potential of current models in complex interactive economies. Our contributions are as follows:

\begin{itemize}
    \item \textbf{Infinite-Horizon Planning Evaluation.} We introduce an open, generalizable framework, \themodel, for assessing LLM agents under continuous, non-episodic interaction, where a compact action space is coupled with an unbounded temporal horizon. This design isolates long-term strategic coherence, stability, and cumulative optimization as first-class evaluation targets, rather than short-horizon task completion.
    \item \textbf{Utility-Guided Economic Assessment.} We select three widely-existed economic environments as testbed: \textit{Vending}, \textit{Freelance} and \textit{Operation}, and establish an outcome-oriented evaluation paradigm grounded in economic returns, moving to measure agent behavior by its tangible impact in market settings.
    \item \textbf{A Multi-Dimensional Empirical Analysis.} We conduct a rigorous evaluation of state-of-the-art LLMs, uncovering a critical performance gap where no single model dominates across all economic scenarios. Through eight meticulously designed diagnostic studies: covering context window lengths, memory modules, and more, we provide deep insights into the current bottlenecks of long-horizon planning in interactive economies.
\end{itemize}

\section{Related Work}
\begin{figure}[t]
    \centering
    \includegraphics[width=\linewidth]{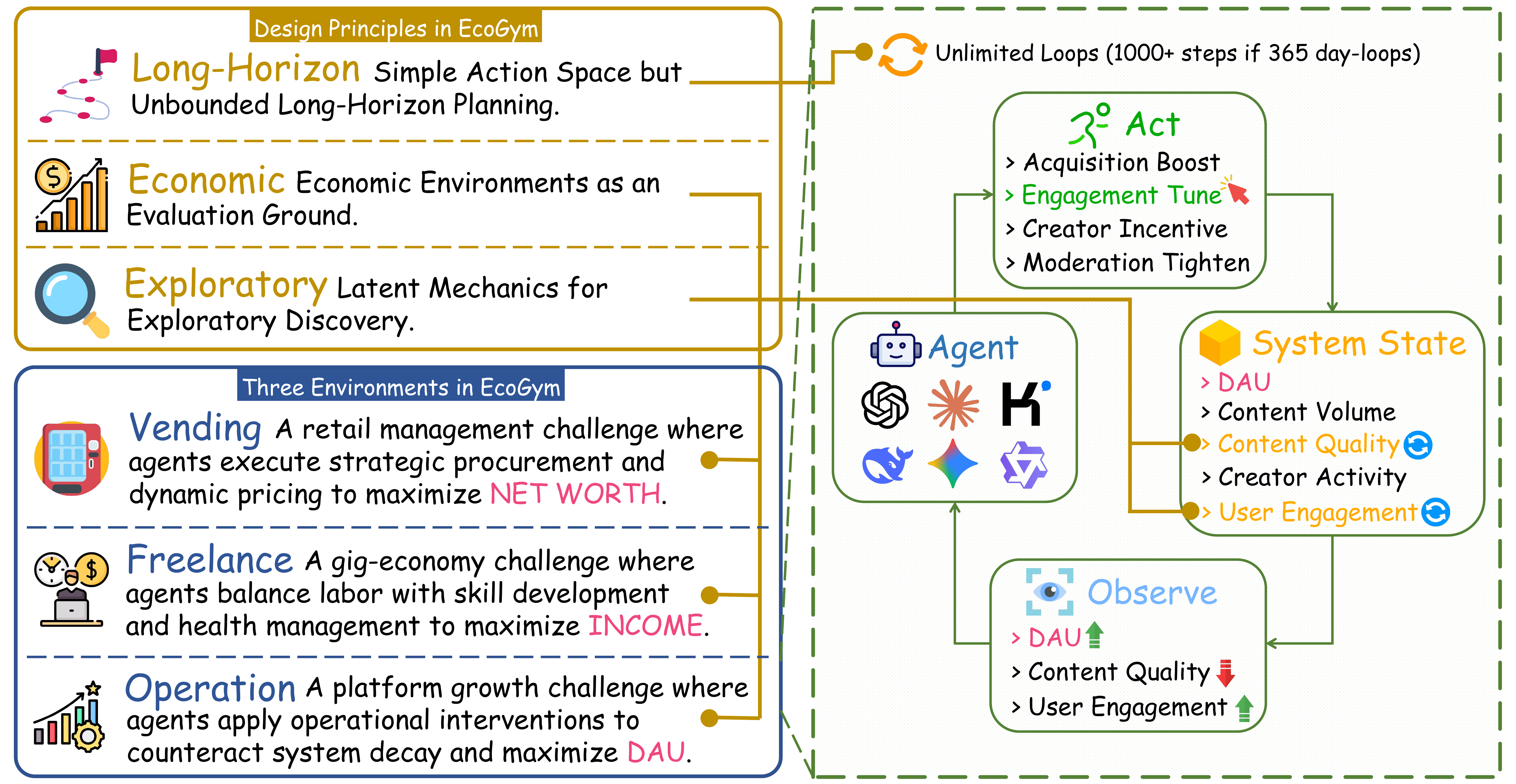}
    \caption{Design Principles (upper left) and three Environments in \themodel (lower left), and detailed description for \textit{Vending} environment (right). We marked how our designs reflect the principles by golden leader line.} \label{fig:overview}
\end{figure}

\noindent \textbf{Long-horizon Planning Evaluation.} Planning is widely recognized as a core defining capability of LLM-powered agents~\citep{zhang2025landscapeagenticreinforcementlearning} and has been a primary focus of existing benchmarks. Across diverse domains, including embodied environments (e.g., ALFWorld~\citep{shridhar2021alfworldaligningtextembodied}, SciWorld~\citep{wang2022scienceworldagentsmarter5th}),
GUI navigation (e.g., AndroidWorld~\citep{rawles2025androidworlddynamicbenchmarkingenvironment}, WebArena~\citep{zhou2024webarenarealisticwebenvironment}, Mobile-Env~\citep{zhang2024mobileenvbuildingqualifiedevaluation}), autonomous driving (e.g., PCA-Bench~\citep{chen2024pcabenchevaluatingmultimodallarge}, MetaAD~\citep{jiang2025alphadriveunleashingpowervlms}), 
and deep research tasks (e.g., xBench~\citep{chen2025xbenchtrackingagentsproductivity}, BrowseComp~\citep{wei2025browsecompsimplechallengingbenchmark}), agents are subjected to stringent and fine-grained evaluations of long-horizon planning competence. 
While these environments and benchmarks explicitly or implicitly assess planning within their respective task domains, recent research has increasingly shifted toward evaluating agent planning anchoring it in tangible practical utility rather than purely virtual rewards~\citep{chen2025xbenchtrackingagentsproductivity}. 
Representative efforts include OpenAI’s GDPval~\citep{patwardhan2025gdpval}, which measures agent performance on economically meaningful tasks spanning major sectors of the U.S. economy, and Vending Bench v1\&v2~\citep{backlund2025vending,vendingbench2}, which probes the ability of LLM-based agents to operate simple yet long-running commercial processes, exemplified by autonomous vending machine management.
In contrast, our \themodel is not confined to narrow, scenario-specific testbeds: it supports multi-scenario, infinite-horizon interaction under a unified framework and is released as a fully open-source platform, enabling transparent and community-driven evaluation of agent planning in economic environments.

\noindent \textbf{Agents in Economic Simulations.} Recent LLM-driven economic agents can be broadly organized into a compact taxonomy along the axes of \emph{level of economic abstraction}: \ding{110} \textbf{Micro-Economic Execution Agents}, which models LLMs as autonomous decision-makers at the level of individual transactions, budget allocation, and survival-oriented objectives~\citep{li2025quantagentsmultiagentfinancialsimulated,gao2023largelanguagemodelsempowered}. {Generative Agents}~\citep{park2023generative}, as an early attempt,  introduce persistent memory and planning to sustain long-horizon social and economic interactions. 
{Vending Bench v1 \& v2}~\citep{backlund2025vending,vendingbench2} formalize economic viability and bankruptcy as quantitative evaluation signals. {HeroBench}~\citep{anokhin2025herobench} further emphasizes competitive dynamics and resource scarcity in adversarial economic regimes; \ding{110} \textbf{Macro-Economic Policy and Population Simulators} elevates LLM agents to population-scale actors for studying aggregate behavior, institutional design, and policy-induced equilibria, 
as implemented in EconAgent~\citep{li2024econagent}, FCLAgent~\citep{hashimoto2025agentbasedsimulationfinancialmarket}, StockSim~\citep{papadakis2025stocksimdualmodeorderlevelsimulator}. 

\section{\themodel}
\subsection{Design Principles}
\label{sec:design_philosophy}

As shown in Fig~\ref{fig:overview}, \themodel is designed based on three core principles aimed at rigorously evaluating LLM agents in complex and continuous environments.

\noindent \textbf{Principle 1: Simple Action Space but Unbounded Long-Horizon Planning.} We combine a compact action space with an infinite horizon. Unlike benchmarks that rely on complex, high-dimensional action spaces, our environments restrict the agent to a small set of discrete actions (typically 4-5 primitives). However, the interaction horizon is effectively infinite. This design choice shifts the evaluation focus to long-term strategic planning. Agents must continuously make decisions to optimize a cumulative objective (e.g., maximizing DAU through perpetual controlling several factors in the \textit{Operation} scenario) without a predefined termination state. This allows us to isolate and measure the agent's capacity for stability and optimization over ultra-long contexts.

\noindent \textbf{Principle 2: Economic Environments as an Evaluation Ground.} We ground the evaluation in daily economic activities to assess decision-making in market environments. Our benchmark investigates how agents influence and adapt to economic dynamics. We select three distinct scenarios to evaluate the agent's ability to navigate resource allocation, labor management, and operational efficiency. This approach moves beyond code generation or simple reasoning tasks to assess how LLMs function as economic actors within a system.

\noindent \textbf{Principle 3: Latent Mechanics for Exploratory Discovery.} We incorporate latent environmental mechanics to enforce active exploration. The explicit rules governing the system's underlying dynamics are not disclosed in the prompt to the agents while requiring exploration by themselves. For instance, in the \textit{Operation} scenario, the mathematical relationship between content quality or user engagement and DAU is hidden. The agent must deduce them through interaction and feedback. This design necessitates that agents transition from passive execution to active hypothesis testing and causal discovery to maximize their utility.

\subsection{Implementation}
\label{sec:implementation}

We construct 3 distinct environments in \themodel: \textit{Vending}, \textit{Freelance} and \textit{Operation}, which are widely-existed in our economic activities. Despite their domain-specific differences, all environments are unified under a decision-making process. 
% This framework enforces three structural invariants: (1) \textit{Budgeted Action Space ($N_{max}$)}, where a strict daily action cap necessitates strategic prioritization; (2) \textit{Standardized Interaction Interface}, ensuring all interactions return structured JSON observations for consistent policy updates; and (3) \textit{Unified State Persistence}, using a global registry to ensure long-horizon consistency across simulation epochs.

\noindent \textbf{General Task Formulation.} Agent task is formulated as a decision-making process under partial observability. Formally, the interaction is modeled as a tuple $\langle \mathcal{S}, \mathcal{A}, \mathcal{O}, \mathcal{T}, \mathcal{G} \rangle$. At each time step $t$, the agent receives a structured observation $o_t \in \mathcal{O}$ derived from the latent environmental state $s_t \in \mathcal{S}$. Guided by a semantic Goal $\mathcal{G}$ (e.g., Net Worth maximization), the agent selects an action $a_t \in \mathcal{A}$ from a discrete set. The environment then transitions to $s_{t+1}$ according to the dynamic transition function $\mathcal{T}(s_{t+1}|s_t, a_t)$.

Below, we present the specific dynamics for each scenario. Detailed mathematical formulations of state transitions, termination conditions, and data generation pipelines are provided in Appendix~\ref{app:env_dynamics}. For specific input arguments and observation structures for each action, please refer to the complete Action Schema in Appendix~\ref{app:action_schema}.

% =========================================================================================
% ENVIRONMENT I: VENDING BENCH
% =========================================================================================

\subsubsection{Environment I: Vending}
\label{sec:vending_bench}

This environment is a retail business scenario where agent acts as a sole proprietor aiming to maximize net worth through strategic procurement and dynamic pricing.

\noindent \textbf{Goal.} The objective is to maximize \textit{Net Worth}, calculated as the sum of liquid cash, the wholesale value of on-hand inventory , and the value of pending pre-paid orders.

\noindent \textbf{System State Space.} The global state $S_t = \{M_t, D_t, \mathbf{Q}_t, \mathbf{W}_t, \mathbf{P}_t, \mathbf{O}_t, \mathcal{H}_t, \mathbf{\Theta}_{market}\}$ tracks liquid cash ($M_t$), current day ($D_t$), inventory levels ($\mathbf{Q}_t$), wholesale costs ($\mathbf{W}_t$), retail prices ($\mathbf{P}_t$), pending orders ($\mathbf{O}_t$). Crucially, $\mathbf{\Theta}_{market}$ represents the hidden latent market parameters (e.g., seasonality $\phi$, elasticity $\eta$) excluded from the agent's observation.

\noindent \textbf{Action Space.} The agent operates within a discrete action space constrained by a maximum of daily actions. The action set includes: (1) \textit{Market Exploration}, which queries the environment to discover new products and wholesale costs; (2) \textit{Inventory Procurement}, which submits purchase orders with a fixed lead time; and (3) \textit{Price Adjustment}, for updating the retail price to modulate consumer demand.

\noindent \textbf{Observation Space.} The agent perceives the environment through two channels: (1) \textit{Action Response}, immediate feedback from exploration/pricing; and (2) \textit{Daily Report}, a global financial statement returned after daily transition.

\noindent \textbf{State Transition.} The state evolution proceeds in two phases (see Appendix~\ref{app:vending_dynamics} for details): (1) \textit{Demand Determination}, where realized sales are calculated based on an Elastic Logit Model driven by hidden seasonality curves and price sensitivity; and (2) \textit{Logistics Settlement}, where pending orders are added to inventory only when the delivery date matches the current day.

\noindent \textbf{Data Collection.} The product information is constructed by querying Perplexity\footnote{\url{https://www.perplexity.ai/}} followed \citep{vendingbench2}. Hidden market physics (seasonality, elasticity) are synthesized via LLMs to create merchandise logic.

% =========================================================================================
% ENVIRONMENT II: FREELANCE BENCH
% =========================================================================================

\subsubsection{Environment II: Freelance}
\label{sec:freelance_bench}

This environment is a gig-economy scenario where the agent acts as a freelancer aiming to maximize income while avoiding burnout.

\noindent \textbf{Goal.} The objective is to maximize \textit{Income}, which is the sum of money the agent earned during the labour activities.

\noindent \textbf{System State Space.} The global state $S_t = \{M_t, E_t, St_t, \mathbf{Sk}_t, \mathcal{T}_t, \tau_{burnout}\}$ tracks Money ($M_t$), Energy ($E_t$), Stress ($St_t$), Skills ($\mathbf{Sk}_t$), and the Task Pool ($\mathcal{T}_t$). $\tau_{burnout}$ is the hidden physiological threshold for burnout.

\noindent \textbf{Action Space.} The agent interacts via a discrete action set to manage labor and health. Actions include: (1) \textit{Exploration}, executing a task discovery via free or paid sourcing; (2) \textit{Labor Execution}, solving tasks where energy consumed; (3) \textit{Settlement}, submitting work to an LLM-based Auditor for verification and payment; and (4) \textit{Wellness}, performing restorative actions to reduce stress, incurring immediate financial costs.

\noindent \textbf{Observation Space.} Observations are context-dependent: (1) \textit{Market View}, a standardized Job Board detailing task difficulty and pay; (2) \textit{Execution Feedback}, an Auditor Log revealing payment approval or penalties; and (3) \textit{Somatic Status}, the agent's physiological state to monitor burnout risk.

\noindent \textbf{State Transition.} The evolution involves a rigorous feedback loop (details in Appendix~\ref{app:freelance_dynamics}): (1) \textit{Metabolic Settlement}, where subsistence costs are deducted; (2) \textit{Physiological Feedback}, where success improves skills while failure spikes stress (triggering a ``death spiral'' if stress exceeds a critical threshold); and (3) \textit{Market Evolution}, where unclaimed tasks decay or expire.

\noindent \textbf{Data Collection.} To construct a diverse task pool, we engineered a rigorous data collection pipeline. We first aggregated authoritative cross-domain datasets, including software development (LiveCodeBench~\citep{jain2024livecodebench}, SWE-bench~\citep{jimenez2023swe}), financial (FinQA~\citep{chen2021finqa}, BizFinBench~\citep{lu2025bizfinbench}), STEM (GSM8K~\citep{cobbe2021training}, SciQ~\citep{pedersen2020sciq}), and Legal/Admin (LEXam~\citep{fan2025lexam}, Pile-of-law~\citep{henderson2022pile}). Raw data then undergoes difficulty filtering, where a lightweight model filters out trivial samples to retain only cognitively challenging instances. Subsequently, metadata enters a strategy router \& mutation phase: coding tasks undergo ``Scenario Injection'', wrapping core logic in business contexts via LLM rewriting, while quantitative tasks undergo ``Logic Mutation'', refactoring numerical values and variables to prevent data contamination or memorization. Finally, all tasks must pass a solvability check via both LLM and human check before injection into the database. 

% =========================================================================================
% ENVIRONMENT III: OPERATION BENCH
% =========================================================================================

\subsubsection{Environment III: Operation}
\label{sec:operation_bench}

This environment is a digital content platform, agent acts as a Platform Operator optimizing Daily Active Users (DAU).

\noindent \textbf{Goal.} The objective is to maximize Average DAU ($DAU_{avg}$). The system is characterized by Zero-Attractor dynamics: without intervention, user activity naturally decays to zero.

\noindent \textbf{System State Space.} The state $S_t = \{DAU_t, Vol_t, Qual_t, Act_t, Eng_t, \mathbf{\Phi}_{sys}\}$ tracks Users ($DAU_t$), Content ($Vol_t$), Quality ($Qual_t$), Activity ($Act_t$), and Engagement ($Eng_t$). $\mathbf{\Phi}_{sys}$ contains hidden system coefficients governing decay and noise.

\noindent \textbf{Action Space.} The agent performs atomic interventions ($N_{max}=1$) with stochastic outcomes: (1) \textit{Acquisition Boost} allocates budget to acquire new users; (2) \textit{Engagement Tune} increases stimulation, boosting retention but degrading quality; (3) \textit{Creator Incentive} subsidizes creators to boost production; and (4) \textit{Moderation Tighten} improves quality but suppresses creator activity.

\noindent \textbf{Observation Space.} The observation provides immediate feedback by returning the updated system state following the execution of an action, allowing the agent to directly monitor the post-intervention status.

\noindent \textbf{State Transition.} The evolution models a non-linear system driven by three coupled sub-processes (equations in Appendix~\ref{app:operation_dynamics}): (1) \textit{User Retention}, affected by content scale and quality; (2) \textit{Supply Production}, amplified by creator incentives; and (3) \textit{Quality Entropy}, which naturally decays unless actively governed.
\subsection{Statistics}
\label{sec:env_stats}

We summarize key statistics in Table~\ref{tab:env_stats}. A distinguishing feature of \themodel is the Long-Horizon characteristic: unlike traditional agent benchmarks that typically involve short interaction sequences, our environments can impose an unbounded horizon (daily budget multiply total days), requiring agents to maintain strategic coherence over thousands of decision steps. Also, \textit{Vending} and \textit{Freelance} challenges the agent with a high-dimensional state space, demanding the management of inventory levels across hundreds of SKUs or thousands of tasks.

\begin{center}
\begin{minipage}{0.86\linewidth}
\captionsetup{type=table,hypcap=false}
\caption{\textbf{Key Statistics of \themodel Environments.} We compare the daily decision budget, action diversity, and the scale of the underlying data universe.}
\label{tab:env_stats}
\vspace{-5pt}
\centering
\begingroup
\footnotesize
\setlength{\tabcolsep}{4pt}
\renewcommand{\arraystretch}{0.98}
\begin{tabular}{lccc}
\toprule
\textbf{Environment} & \textbf{Daily Budget} & \textbf{Action Types} & \textbf{Data Scale} \\
\midrule
\textit{Vending} & 4 & 4 & 600+ SKUs, 37 Categories \\
\textit{Freelance} & 5 & 5 & 8 Datasets, 5k+ Tasks \\
\textit{Operation} & 1 & 4 & Continuous Param. Space \\
\bottomrule
\end{tabular}
\endgroup
\end{minipage}
\vspace{-8pt}
\end{center}

\begin{table}[t]
\caption{\textbf{Performance comparison on \themodel.} Best and second best results are indicated by \textbf{bold} and \underline{underlined}.}
\label{tab:main_results}
\begin{center}
% \begin{small}
\resizebox{0.85\textwidth}{!}{
% \begin{tabular}{lccc}
% \toprule
%     Model & Vending (Net Worth) $\uparrow$ & Freelance (Income) $\uparrow$ & Operation (DAU) $\uparrow$ \\
% \midrule
%     Claude-Sonnet-4.5 & 1816.62 & 243.18 $\pm$ 307.72 & \textbf{1595.00 $\pm$ 57.78} \\
%     DeepSeek-v3.2     & 2775.94 & 283.10 $\pm$ 405.80 & 1210.33 $\pm$ 178.46 \\
%     Gemini-3-Flash    & \underline{5675.21} & 1577.10 $\pm$ 483.44 & 1211.20 $\pm$ 167.00 \\
%     Gemini-3-Pro      & \textbf{11274.73} & \underline{2548.78 $\pm$ 238.14} & 1266.80 $\pm$ 168.83 \\
%     GLM-4.7           & 938.64 & 910.39 $\pm$ 55.40 & 1198.33 $\pm$ 77.22 \\
%     GLM-5           & - & 1046.21 $\pm$ 190.78 & 1271.67 $\pm$ 159.80 \\
%     GPT-5-Mini        & 474.63 & \textbf{2990.72 $\pm$ 298.17} & 1215.00 $\pm$ 54.97 \\
%     GPT-5.2           & 1062.09 & 1434.26 $\pm$ 148.14 & 1095.60 $\pm$ 58.23 \\
%     Grok-4.1-Fast     & 1745.69 & 293.12 $\pm$ 418.23 & \underline{1385.67 $\pm$ 76.77} \\
%     % Kimi-k2           & 942.49 & 0.00 & 1109.05 \\
%     Kimi-k2.6           & - & 1199.12 $\pm$ 54.95 & 1191.67 $\pm$ 162.17 \\
%     Qwen3-235b-A22b   & 674.45 & 795.13 $\pm$ 31.85 & 1130.67 $\pm$ 37.72 \\
%     Qwen3.5-397b-A17b   & - & 1029.23 $\pm$ 44.93 & 1081.33 $\pm$ 73.14 \\
%     MiniMax-M2.1      & 1871.80 & 494.71 $\pm$ 336.79 & 989.00 $\pm$ 64.36 \\
%     MiniMax-M2.7      & - & 393.11 $\pm$ 329.10 & 1247.67 $\pm$ 34.87 \\
% \bottomrule
% \end{tabular}
\begin{tabular}{lccc}
\toprule
    Model & Vending (Net Worth) $\uparrow$ & Freelance (Income) $\uparrow$ & Operation (DAU) $\uparrow$ \\
\midrule
    % Claude-Sonnet-4.5 & 1843.62 $\pm$ 329.46 & 241.75 $\pm$ 374.41 & \textbf{1595.00 $\pm$ 70.77} \\
    % Gemini-3-Pro & \textbf{14026.25 $\pm$ 3221.61} & \underline{2696.58 $\pm$ 229.97} & 1266.80 $\pm$ 188.76 \\
    % Gemini-3-Flash & \underline{5730.51 $\pm$ 552.50} & 1815.95 $\pm$ 569.40 & 1301.40 $\pm$ 50.12 \\
    % GLM-4.7 & 458.85 $\pm$ 59.76 & 899.62 $\pm$ 70.13 & 1198.33 $\pm$ 94.57 \\
    % GLM-5 & 933.21 $\pm$ 306.64 & 1022.21 $\pm$ 294.32 & 1271.67 $\pm$ 195.71 \\
    % GPT-5-Mini & 474.63 $\pm$ 140.52 & \textbf{2992.72 $\pm$ 332.87} & 1215.00 $\pm$ 61.46 \\
    % GPT-5.2 & 1062.09 $\pm$ 269.66 & 1434.26 $\pm$ 165.63 & 1095.60 $\pm$ 65.11 \\
    % Grok-4.1-Fast & 2092.48 $\pm$ 594.61 & 296.45 $\pm$ 509.38 & \underline{1385.67 $\pm$ 94.03} \\
    % Kimi-k2 & 925.80 $\pm$ 309.61 & 659.37 $\pm$ nan & 1145.00 $\pm$ nan \\
    % Kimi-K2.6 & 1094.96 $\pm$ 243.82 & 1072.06 $\pm$ 187.96 & 1191.67 $\pm$ 198.61 \\
    % MiniMax-M2.1 & 277.13 $\pm$ 74.93 & 488.10 $\pm$ 412.63 & 1050.00 $\pm$ 161.17 \\
    % Qwen3-235b-A22b & 370.29 $\pm$ 47.05 & 795.13 $\pm$ 39.01 & 1130.67 $\pm$ 46.20 \\
    % Qwen3.5-397b-A17b & 1069.43 $\pm$ 493.86 & 1029.23 $\pm$ 55.03 & 1081.33 $\pm$ 89.58 \\
    % DeepSeek-v3.2 & 1811.99 $\pm$ 518.72 & 286.43 $\pm$ 494.14 & 1210.33 $\pm$ 218.57 \\
    % GLM-5.1 & 1586.22 $\pm$ 251.04 & \color{red}{1000.00 $\pm$ 0.00} & 1189.00 $\pm$ 187.29 \\
    % MiniMax-M2.7 & 1712.73 $\pm$ 290.27 & 393.11 $\pm$ 403.06 & 1247.67 $\pm$ 42.71 \\
    Claude-Sonnet-4.5 & 1843.62 $\pm$ 329.46 & 241.75 $\pm$ 374.41 & \textbf{1595.00 $\pm$ 70.77} \\
    Gemini-3-Pro & \textbf{11272.47 $\pm$ 1037.27} & \textbf{2696.58 $\pm$ 229.97} & \underline{1396.00 $\pm$ 53.73} \\
    Gemini-3-Flash & \underline{5675.65 $\pm$ 633.67} & \underline{1815.95 $\pm$ 569.40} & 1194.00 $\pm$ 55.38 \\
    GPT-5.2 & 1062.09 $\pm$ 269.66 & 1434.26 $\pm$ 165.63 & 1095.60 $\pm$ 65.11 \\
    Kimi-K2.6 & 1094.96 $\pm$ 243.82 & 1072.06 $\pm$ 187.96 & 1191.67 $\pm$ 198.61 \\
    Qwen3.5-397b-A17b & 1069.43 $\pm$ 493.86 & 1029.23 $\pm$ 55.03 & 1081.33 $\pm$ 89.58 \\
    DeepSeek-v3.2 & 2125.18 $\pm$ 411.28 & 286.43 $\pm$ 494.14 & 1210.33 $\pm$ 218.57 \\
    GLM-5 & 933.21 $\pm$ 306.64 & 1022.21 $\pm$ 294.32 & 1271.67 $\pm$ 195.71 \\
    MiniMax-M2.7 & 1712.73 $\pm$ 290.27 & 393.11 $\pm$ 403.06 & 1247.67 $\pm$ 42.71 \\
\bottomrule
\end{tabular}
}
% \end{small}
\end{center}
\vspace{-10pt}
\end{table}

\section{Experiments}

We conduct comprehensive experiments across a diverse spectrum of LLMs, encompassing flagship models from different organizations. To go beyond standard metrics and thoroughly characterize agent performance—particularly within long-horizon tasks, we design a rigorous suite of analytical experiments or case study to provide insights for the optimization of models or harnesses. These evaluations focus on 8 critical dimensions: stochastic system stability, context length, failure modes, the temporal evolution of agent behaviors, comparison with human baselines, additional memory modules, thinking with action and environment complexity.

\subsection{Setup}

\noindent \textbf{Models.} We benchmark a diverse set of models, spanning proprietary and open-weights LLMs as follows: \ding{110} \textbf{Proprietary:} GPT-5.2, Gemini-3-Pro, Gemini-3-Flash, Claude-Sonnet-4.5, Kimi-k2.6, MiniMax-M2.7. \ding{110} \textbf{Open-Weights:} Qwen3-235b-A22b, DeepSeek-v3.2, GLM-5. Refer Appendix~\ref{ap:api_info} for detailed version information about the model. Generation parameters are standardized (Temperature=1.0, Top-p=0.95) across all trials to ensure fair comparison. The maximum days are set to 365 days equivalent to one year. The context window is restricted to a sliding window of the most recent 128 steps.

\subsection{Experimental Results}
\label{sub:exp_main}

\noindent \textbf{Main Results.} Table~\ref{tab:main_results} presents the comparative performance across three environments. For \textit{Vending}, we conduct experiments for five times and report the average due to its high variance across runs, while \textit{Freelance} and \textit{Operation} are for three times to reduce costs, as our pilot studies indicated relatively lower variance for these tasks. Detailed analysis on variance is in later paragraphs. In \textit{Vending}, Gemini-3 series demonstrates dominant asset appreciation compared to others. Shifting to \textit{Freelance}, the Gemini series maintains its lead, though the performance margin over its competitors is notably narrower. In \textit{Operation}, Claude-Sonnet-4.5 ranks first. Crucially, these results indicate that no single model maintains a dominant lead on \themodel. This lack of a universal winner underscores the challenging nature of the benchmark and highlights significant room for future improvement.

% \begin{figure}[t]
%     \centering
    
%     \begin{subfigure}{0.32\linewidth}
%         \centering
%         \includegraphics[width=\linewidth]{figures/stability/stability_vending_Gemini-3-Pro.pdf}
%     \end{subfigure}
%     \begin{subfigure}{0.32\linewidth}
%         \centering
%         \includegraphics[width=\linewidth]{figures/stability/stability_freelance_Gemini-3-Pro.pdf}
%     \end{subfigure}
%     \begin{subfigure}{0.32\linewidth}
%         \centering
%         \includegraphics[width=\linewidth]{figures/stability/stability_operation_Gemini-3-Pro.pdf}
%     \end{subfigure}
    
%     \caption{Stochastic stability analysis of Gemini-3-Pro on \textit{Vending} (left), \textit{Freelance} (middle) and \textit{Operation} (right).}
%     \label{fig:stability_Gemini-3-Pro}
% \end{figure}

\textbf{Stochastic Stability and Variance Analysis.}\label{sub:exp_stability} Model performance in long-horizon environments is prone to inherent instability. To quantify this stochasticity, we conducted independent trials across all three scenarios. 
%The stability profile for Gemini-3-Pro is illustrated in Figure~\ref{fig:stability_Gemini-3-Pro}, while comprehensive variance plots for the remaining model suite are provided in Appendix~\ref{app:stability}. 
Our analysis reveals environmental disparities within \themodel: the \textit{Vending} environment exhibits high performance variance (the same as the original Vending bench), whereas agents in \textit{Freelance} and \textit{Operation} demonstrate relatively stable trajectories. Consequently, to ensure robust evaluation, the main results in Section~\ref{sub:exp_main} report the average of five runs for \textit{Vending}, compared to three runs reporting for the more deterministic \textit{Freelance} and \textit{Operation} tasks.

\begin{figure}[t]
    \centering
    \begin{minipage}[t]{0.49\linewidth}
        \centering
        \includegraphics[width=\linewidth]{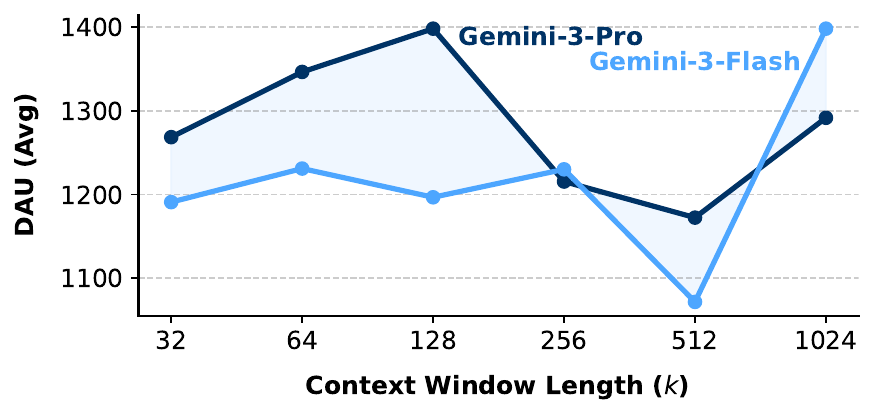}
        \caption{\textbf{Impact of context window length on \textit{Operation}.} Gemini-3-Flash and Gemini-3-Pro are compared across lengths from 32 to 1024.}
        \label{fig:context_window}
    \end{minipage}
    \hfill
    \begin{minipage}[t]{0.49\linewidth}
        \centering
        \includegraphics[width=\linewidth]{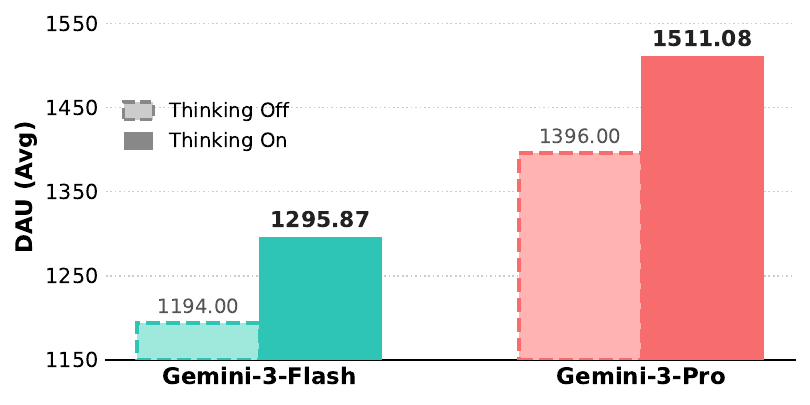}
        \caption{\textbf{Effect of Thinking mode on \textit{Operation}.} We compare disabled (\textit{Off}) vs. enabled (\textit{On}).}
        \label{fig:thinking}
    \end{minipage}
\end{figure}

\noindent \textbf{Impact of Context Window Length.} \label{sub:exp_context} A pivotal question is whether extending the context window beyond the default setting ($k=128$) yields performance gains. To investigate this, we evaluate Gemini-3-Flash and Gemini-3-Pro within the \textit{Operation} environment across context lengths ranging from $k=32$ to $k=1024$, with results summarized in Figure~\ref{fig:context_window}. Notably, we find that expanding the context window does not yield consistent performance gains. Gemini-3-Flash demonstrates a volatile trajectory: it shows initial improvements from $k=32$ followed by a decline, yet significantly rebounds at $k=1024$ to rival the optimal performance of Gemini-3-Pro ($k=128$). In stark contrast, Gemini-3-Pro exhibits a divergent behavior where performance peaks at $k=128$ and degrades progressively as the window extends to $k=1024$. These disparate outcomes underscore the significant instability inherent in current models when processing super long contexts.

\noindent \textbf{Failure Modes Analysis.}
\label{sub:exp_failure}
Unlike traditional benchmarks anchored by static gold-standard answers, \themodel presents an open-ended optimization challenge where success is defined by relative cumulative rewards. To isolate the behavioral drivers of performance, we conducted a differential trajectory analysis between Top-2 models across three scenarios inspired original vending bench paper \citep{backlund2025vending}, employing a human-in-the-loop workflow assisted by LLMs. Our analysis reveals that performance gaps stem primarily from two distinct capabilities: (1) \textit{Strategic Prioritization}. Superior models demonstrate better alignment with the underlying reward mechanism. For instance, in the \textit{Operation} task, the leading Claude-Sonnet-4.5 prioritized scale (generating 643 items with 0.566 quality) over the runner-up's focus on refinement (326 items with 0.762 quality), correctly identifying quantity as the dominant variable for maximizing yield. (2) \textit{Execution Efficiency}. Top models exhibit significantly higher action utility. In \textit{Vending}, Gemini-3-Pro actively leveraged daily allowances for market exploration, whereas the second-place Gemini-3-Flash frequently defaulted to passive waiting. Similarly, in \textit{Freelance}, Gemini-3-Pro demonstrated precise state tracking with negligible invalid actions, in stark contrast to Gemini-3-Flash, which suffered from redundant loops (e.g., repeated task queries), indicating deficiencies in long-context state maintenance.

\begin{figure}[t]
    \centering
    
    \begin{subfigure}{0.32\linewidth}
        \centering
        \includegraphics[width=\linewidth]{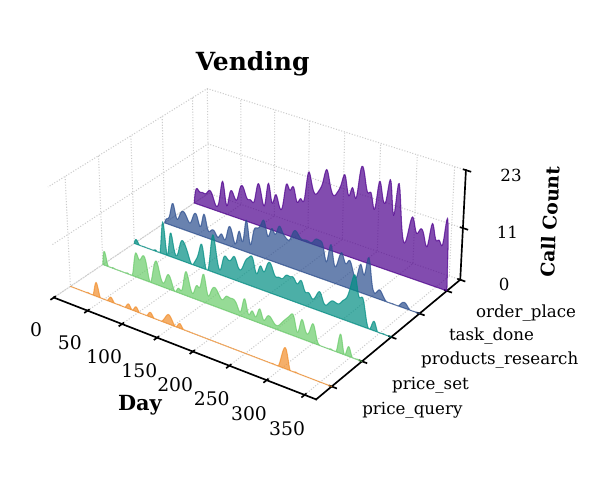}
    \end{subfigure}
    \begin{subfigure}{0.32\linewidth}
        \centering
        \includegraphics[width=\linewidth]{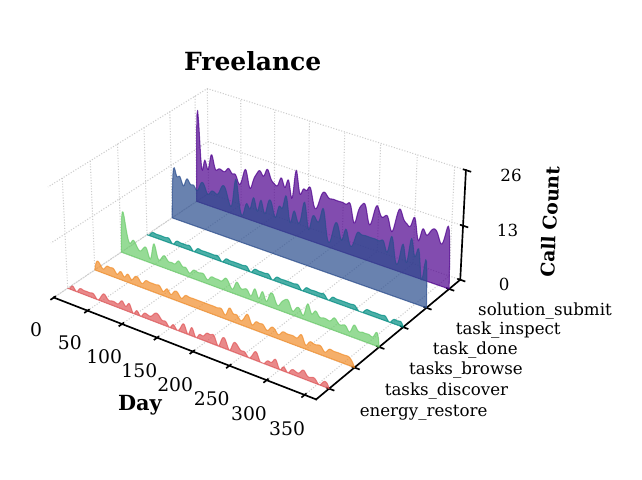}
    \end{subfigure}
    \begin{subfigure}{0.32\linewidth}
        \centering
        \includegraphics[width=\linewidth]{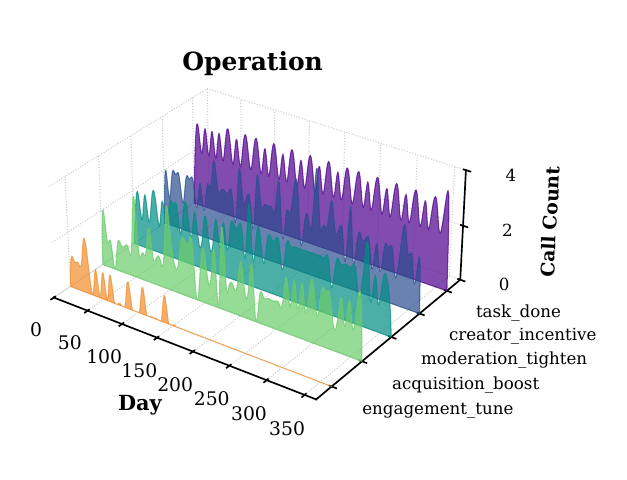}
    \end{subfigure}
    
    \caption{Temporal evolution of action frequencies for Gemini-3-Pro in \textit{Vending} (left), \textit{Freelance} (middle) and \textit{Operation} (right).}
    \label{fig:tool_Gemini-3-Pro}
\end{figure}

\noindent \textbf{Temporal Evolution of Agent Behavioral Patterns.}
\label{sub:exp_tools} 
As illustrated in Figure~\ref{fig:tool_Gemini-3-Pro}, this methodology filters stochastic noise to reveal macro-scale cognitive phase transitions for Gemini-3-Pro across diverse environments, with other model's results in Appendix~\ref{app:tool_use}. In \textit{Vending}, the agent exhibits a precise ``cold-start versus steady-state'' dichotomy, where initial intense exploration via \texttt{products\_research} and \texttt{price\_set} rapidly decays into a stable, cyclical \texttt{order\_place} replenishment loop, indicating an implicit optimization of cognitive overhead. Conversely, a dynamic homeostatic mechanism is revealed in \textit{Freelance}; the agent establishes a rhythmic oscillation between \texttt{task\_inspect}, \texttt{solution\_submit}, and \texttt{energy\_restore}, balancing immediate economic utility with physiological maintenance while sustaining background \texttt{tasks\_discover} activity. Finally, in \textit{Operation}, agent sequentially shifts focus from \texttt{acquisition\_boost} to \texttt{moderation\_tighten} and ultimately \texttt{creator\_incentive} (retention), evidencing the model's capacity for state-dependent strategic planning rather than myopic instruction following.

\noindent \textbf{Comparative Analysis against Human Performance.} \label{sub:exp_human} To establish a rigorous baseline for assessing LLM capabilities in economic tasks, we recruited 3 human experts to perform evaluations within \themodel. To facilitate human interaction, we developed a dedicated GUI, as visualized in Appendix~\ref{app:gui}. Given that tasks such as \textit{Vending} and \textit{Freelance} involve extensive temporal horizons—typically requiring 1,500 to 2,000 interaction steps and spanning several hours—maintaining consistent human attention proved to be a significant challenge. Consequently, we restricted the human evaluation in \textit{Operation} environment. Prior to the formal evaluation, experts were granted a 15-minute session to familiarize themselves with the environment; notably, no information regarding the underlying economic mechanisms was disclosed to them. Human experts took approximately 45 minutes to complete a single episode, achieving an average DAU of 1,404. Remarkably, Claude-Sonnet-4.5 surpassed this human baseline. This result demonstrates that current SOTA LLMs have potential to achieve super-human performance in specific long-horizon economic planning scenarios, highlighting their immense potential for complex economic decision-making.

\noindent \textbf{Impact of Additional Memory Modules.} \label{sub:exp_memory}
Augmenting LLM agents with external memory modules represents a promising avenue for mitigating long-horizon context limitations. We implement and evaluate three canonical memory architectures: working, symbolic, and episodic, alongside a commercial solution, Mem0. Implementation details are provided in Appendix~\ref{ap:memory_imple}. As detailed in Table~\ref{tab:memory_ablation}, our results reveal that: (1) while memory integration generally enhances performance, it is not universally beneficial (e.g., Gemini-3-Pro with working memory in \textit{Freelance} shows performance regression); (2) the efficacy of memory mechanisms is highly model-dependent, as Gemini-3-Flash consistently favors Working Memory across all benchmarks, whereas Gemini-3-Pro benefits from alternative architectures; and (3) optimal memory selection exhibits significant task-dependence, evidenced by Gemini-3-Pro requiring distinct memory types across the three different environments. These findings collectively indicate that no single memory paradigm currently holds a dominant position.

\begin{table}[t]
\begin{minipage}[t]{0.59\linewidth}
    \centering
    \captionsetup{font=small,skip=7pt}
    \caption{\textbf{Memory ablation.} $\mathcal{M}_{\text{work}}$, $\mathcal{M}_{\text{sym}}$, $\mathcal{M}_{\text{epi}}$, and $\mathcal{M}_{\text{mem0}}$ denote working, symbolic, episodic, and Mem0 memory. Best and second best results are shown in \textbf{bold} and \underline{underlined}.}
    \label{tab:memory_ablation}
    \begingroup
    \setlength{\tabcolsep}{4pt}
    \renewcommand{\arraystretch}{1.2}
    \resizebox{\linewidth}{!}{
    \begin{tabular}{lccccc}
    \toprule
    Model & None & $+ \mathcal{M}_{\text{work}}$ & $+ \mathcal{M}_{\text{sym}}$ & $+ \mathcal{M}_{\text{epi}}$ & $+ \mathcal{M}_{\text{mem0}}$ \\
    \midrule
    \textit{Gemini-3-Flash} & 5676 & \textbf{10099} & \underline{9661} & 8127 & 7360 \\
    \textit{Gemini-3-Pro}   & 11282 & 13679 & \underline{15567} & \textbf{18939} & 15406 \\
    \bottomrule
    \end{tabular}
    }
    \endgroup
\end{minipage}
\hfill
\begin{minipage}[t]{0.38\linewidth}
    \centering
    \captionsetup{font=small,skip=7pt}
    \caption{\textbf{Performance across task complexities.} Results are reported as mean $\pm$ std; higher is better.}
    \label{tab:complexity}
    \begingroup
    \setlength{\tabcolsep}{4pt}
    \renewcommand{\arraystretch}{1.2}
    \resizebox{\linewidth}{!}{
    \begin{tabular}{lcc}
    \toprule
        Complexity & Gemini-3-Flash & Gemini-3-Pro \\
    \midrule
        Small      & 3512.47 $\pm$ 734.82  & 14089.53 $\pm$ 1243.76 \\
        Medium     & 3889.14 $\pm$ 921.09  & 12187.34 $\pm$ 1787.52 \\
        Large      & 5675.65 $\pm$ 633.67  & 11272.47 $\pm$ 1037.27 \\
    \bottomrule
    \end{tabular}
    }
    \endgroup
\end{minipage}
% \resizebox{\columnwidth}{!}{
% \begin{tabular}{lccc}
% \toprule
%     Memory Component & Net Worth $\uparrow$ & Income $\uparrow$ & Total Revenue $\uparrow$ \\
% \midrule
% \multicolumn{4}{l}{\textit{Gemini-3-Flash}} \\
%     None                          & 5675.21 & 1730.61 & 13705.06 \\
%     + $\mathcal{M}_{\text{work}}$ & \textbf{10099.44} & \textbf{7457.97} & \textbf{24356.81} \\
%     + $\mathcal{M}_{\text{sym}}$  & \underline{9661.01} & \underline{5272.47} & \underline{22568.88} \\
%     + $\mathcal{M}_{\text{epi}}$  & 8127.54 & 5134.62 & 20321.91 \\
%     + $\mathcal{M}_{\text{mem0}}$ & 7360.78 & 3941.56 & 18006.93 \\
% \midrule
% \multicolumn{4}{l}{\textit{Gemini-3-Pro}} \\
%     None                          & 11274.73 & \underline{5246.56} & 26062.45 \\
%     + $\mathcal{M}_{\text{work}}$ & 13679.12 & 4172.43 & \underline{35557.70} \\
%     + $\mathcal{M}_{\text{sym}}$  & \underline{15567.21} & 3381.00 & \textbf{38209.51} \\
%     + $\mathcal{M}_{\text{epi}}$  & \textbf{18939.92} & 4477.13 & 42214.61 \\
%     + $\mathcal{M}_{\text{mem0}}$ & 15406.74 & \textbf{6365.17} & 36744.08 \\
% \bottomrule
% \end{tabular}
% }
\vspace{-10pt}
\end{table}

\noindent \textbf{Impact of Thinking with Action.}\label{exp:thinking} To evaluate the performance for models's\ thinking with action, we assess agents within the Operation environment and generated thinking content are incrementally integrated into the context. As reported in Figure~\ref{fig:thinking}, the activation of the Thinking mode catalyzes a universal performance elevation across both models and all evaluated metrics. For the lightweight Gemini-3-Flash, the introduction of explicit reasoning yields a substantial gain, with DAU increasing from 1194.00 to 1396.00, effectively narrowing the performance gap between the Flash and Pro variants. Similarly, Gemini-3-Pro achieves its highest observed performance with Thinking enabled. These results collectively underscore that maintaining a persistent reasoning chain within the context not only enhances the stability of agentic trajectories but also optimizes the overall success rate in long-horizon tasks, regardless of the model's inherent capacity.

\noindent \textbf{Impact of Environment Complexity.}
\label{sub:exp_complexity} 
To investigate performance on different environment complexity, we curated the \textit{Vending} environment into three complexity tiers by modulating inventory size: Large ($N=37$), Medium ($N=16$), and Small ($N=8$), while preserving the underlying supply-demand dynamics. Theoretically, expanding the inventory raises the potential profit ceiling but simultaneously escalates the cognitive load required for optimal decision-making. As shown in Table~\ref{tab:complexity}, models exhibit divergent scaling behaviors. Gemini-3-Flash demonstrates robust adaptability, with profits scaling positively alongside complexity. In stark contrast, Gemini-3-Pro stagnates, failing to capture additional value in Medium and Large settings despite the broader opportunity space. This inability to exploit expanded state spaces exposes the brittleness of current models when navigating high-dimensional planning tasks within \themodel.

\section{Conclusion}

This work introduces \themodel, addressing the critical void in existing benchmarks regarding the evaluation of agents' long-term economic viability. By moving beyond atomic task success and establishing an infinite-horizon interactive environment, we force agents to navigate the complexities of resource scarcity and stochastic dynamics over sustained periods. Our extensive experiments reveal that while frontier models demonstrate impressive short-term reasoning, they struggle to maintain strategic coherence over long-time decisions. However, we acknowledge that the current framework relies on rule-based abstractions that may not fully capture the unpredictability of irrational human behavior. Furthermore, while this study focuses on individual agent-environment interactions, incorporating multi-agent game-theoretic dynamics remains a vital next step to market pressures. We hope this work guide the community toward developing general-purpose agents that are not only capable of reasoning but are also robust and strategically aligned over the long haul.

\newpage
\section*{Contributions}
\begin{multicols}{2}
\textbf{Core Contributors}
\begin{itemize}
    \item Xavier Hu
    \item Jinxiang Xia
    \item Shengze Xu
    \item Kangqi Song
\end{itemize}
\textbf{Contributors}
\begin{itemize}
    \item Yishuo Yuan
    \item Guibin Zhang
    \item JinCheng Ren
    \item Boyu Feng
    \item Li Lu
    \item Tieyong Zeng
    \item Jiaheng Liu
    \item Minghao Liu
    \item Yuchen Eleanor Jiang
    \item Wei Wang
\end{itemize}

\textbf{Corresponding Authors}
\begin{itemize}
    \item He Zhu
    \item Wangchunshu Zhou
    \item[]
    \item[]
    \item[]
    \item[]
    \item[]
    \item[]
    \item[]
    \item[]
    \item[]
    \item[]
    \item[]
    \item[]
\end{itemize}
\end{multicols}

\clearpage
\bibliographystyle{plainnat}
\bibliography{references}

\appendix
\onecolumn
\section{Model API Detailed Information} \label{ap:api_info}

For proprietary models, we utilize specific version snapshots where available to account for the ``model drift'' often observed in rolling-update API services.
Table~\ref{table:proprietary_api_info} lists the exact version identifiers and the official documentation entries used during our evaluation period (spanning late 2025 to early 2026).

\begin{table}[t]
\caption{Detailed version identifiers and API endpoints for proprietary models.}
\label{table:proprietary_api_info}
\centering
\small
\setlength{\tabcolsep}{35pt}
\begin{tabularx}{1\linewidth}{@{\hspace{15pt}} l X @{}}
\toprule
\textbf{Model Version Identifier} & \textbf{API Entry} \\
\midrule
    GPT-5.2-2025-12-11 & \url{https://platform.openai.com/docs/models/gpt-5.2} \\
    GPT-5-Mini-2025-08-07 & \url{https://platform.openai.com/docs/models/gpt-5-mini} \\
    Gemini-3-Pro-Preview & \url{https://ai.google.dev/gemini-api/docs/gemini-3} \\
    Gemini-3-Flash-Preview & \url{https://ai.google.dev/gemini-api/docs/gemini-3} \\
    Claude-Sonnet-4-5-20250929 & \url{https://platform.claude.com} \\
    Kimi-K2.6 & \url{https://platform.moonshot.cn} \\
    MiniMax-M2.7 & \url{https://platform.minimaxi.com} \\
\bottomrule
\end{tabularx}
\end{table}

Detailed repository information for all evaluated open-weights models is provided in Table~\ref{table:open_weights_info} to facilitate reproducibility and further research.

\begin{table}[t]
\caption{Repository links for open-weights models evaluated in this study.}
\label{table:open_weights_info}
\centering
\small
\setlength{\tabcolsep}{28pt}
\begin{tabularx}{1\linewidth}{@{\hspace{15pt}} l X @{}}
\toprule
\textbf{Model Name} & \textbf{Model Link} \\
\midrule
    Qwen3.5-397b-A17b & \url{https://huggingface.co/Qwen/Qwen3.5-397B-A17B} \\
    DeepSeek-V3.2 & \url{https://hf.co/deepseek-ai/DeepSeek-V3.2} \\
    GLM-5 & \url{https://hf.co/zai-org/GLM-5} \\
\bottomrule
\end{tabularx}
\end{table}

\section{Detailed Environment Dynamics}
\label{app:env_dynamics}

This appendix provides the complete mathematical formulation for the state transitions, termination conditions, and dynamics of the three environments. 

\subsection{Vending}
\label{app:vending_dynamics}

\paragraph{Metric Calculation.}
Net Worth is defined as:
\begin{equation}
    \text{Net Worth} = M_{final} + \mathbf{Q}_{final}^\top \mathbf{W} + \text{Val}(\mathbf{O}_{pending})
\end{equation}
\noindent where $M_{final}$ is the liquid cash, $\mathbf{Q}_{final}^\top \mathbf{W}$ represents the dot product of the inventory vector and wholesale cost vector (total inventory value), and $\text{Val}(\mathbf{O}_{pending})$ estimates the value of pre-paid orders yet to arrive.

\paragraph{Procurement Constraints.}
The Immediate Payment Model enforces strict liquidity:
\begin{equation}
    \text{If } M_t < C_{order}: \text{Reject Transaction}
\end{equation}
\begin{equation}
    \text{Else: } M_t \leftarrow M_t - C_{order}, \quad \mathbf{O}_t \leftarrow \mathbf{O}_t \cup \{order\}
\end{equation}
\noindent where $M_t$ is the current cash balance, $C_{order}$ is the total cost of the purchase order, and $\mathbf{O}_t$ is the set of pending delivery orders.

\paragraph{Termination Condition.}
Failure occurs upon bankruptcy or stagnation:
\begin{equation}
    \text{Fail} \iff (D_t > T_{max}) \lor \left((M_t \le 0) \land (\tau_{no\_sales} \ge \tau_{limit})\right)
\end{equation}
\noindent where $D_t$ is the current day, $\tau_{no\_sales}$ is the count of consecutive days without revenue, and $\tau_{limit}$ is the allowed stagnation threshold.

\paragraph{Demand Simulation (Elastic Logit Model).}
1. \textit{Base Demand:} Follows a hidden seasonality curve:
\begin{equation}
    D_g^{base}(t) = Base \cdot \left(1 + Amp \cdot \sin\left(\frac{2\pi}{T}t + \phi\right)\right)
\end{equation}

2. \textit{Elasticity Adjustment:}
\begin{equation}
    D_g^{total} = D_g^{base}(t) \cdot \left( \frac{\bar{P}_g}{\bar{P}_{ref, g}} \right)^{\eta_{group}}
\end{equation}
\noindent where $\bar{P}_g$ is the average retail price, $\bar{P}_{ref, g}$ is the reference price (typically derived from wholesale cost $\bar{W}_g$), and $\eta_{group} < 0$ is the negative price elasticity coefficient.

3. \textit{Market Share:} Determined via Softmax on utility $U_i$:
\begin{equation}
    U_i = -\beta \frac{P_i}{P_{ref, i}}, \quad s_i = \frac{\exp(U_i) \cdot \mathbb{I}[\mathbf{Q}_t[i] > 0]}{\sum_{j \in g} \exp(U_j) \cdot \mathbb{I}[\mathbf{Q}_t[j] > 0]}
\end{equation}
\noindent where $P_{ref, i}$ is the item-specific reference price and $\beta$ is the price sensitivity factor.

4. \textit{Realization:} $Sold_i = \min(\text{round}(\max(0, s_i D_g^{total} + \mathcal{N}(0, \sigma^2))), \mathbf{Q}_t[i])$.

\subsection{Freelance}
\label{app:freelance_dynamics}

\paragraph{Metric Calculation.}
The primary goal is to maximize \textbf{Income} ($M_{final}$), calculated as:
\begin{equation}
    Income = R_{total} - (C_{daily} + C_{refresh} + C_{relax})
\end{equation}
\noindent The \textbf{Composite Score} ($S_{total}$) aggregates wealth and a comprehensive ``Human Capital'' score to reflect long-term sustainability:
\begin{equation}
    S_{total} = w_1 \cdot M_{final} + w_2 \cdot \left( \lambda_s \overline{\mathbf{Sk}}_{final} + \lambda_e E_{final} + \lambda_r (St_{max} - St_{final}) \right)
\end{equation}
\noindent where the second term rewards high skills ($\overline{\mathbf{Sk}}$), high energy ($E$), and low stress ($St$), aligning with the agent's survival constraints.

\paragraph{Termination Condition.}
\begin{equation}
    \text{Fail} \iff (D_t > T_{max}) \lor (M_t \le 0) \lor (E_t \le 0) \lor (St_t \ge St_{max})
\end{equation}
\noindent where $E_t$ is Energy, $St_t$ is Stress, and $St_{max}$ is the burnout threshold.

\paragraph{Physiological Coupling.}
Energy cost is governed by the Skill-Difficulty coupling law:
\begin{equation}
    \Delta E_{cost} = E_{base} + \alpha D \cdot \max\left(\epsilon_{min}, 1 - \frac{Sk - D}{\lambda}\right)
\end{equation}
\noindent where $\epsilon_{min}$ is the minimum marginal energy cost ensuring non-zero consumption.

The Burnout Mechanism activates if Stress exceeds $St_{crit}$:
\begin{equation}
    P(\text{Fail})_{t+1} \leftarrow P(\text{Fail})_t \cdot \gamma \quad (\text{where } \gamma > 1)
\end{equation}

\paragraph{LLM-as-a-Judge.}
For each test instance, the judge is presented with following information: the task specification, the agent-generated solution, and the ground-truth reference. The model is then tasked with producing a binary assessment (i.e., Correct or Incorrect). To ensure the reliability and alignment of this automated metric, we conducted a rigorous human validation study on a subset of 100 randomly sampled instances. These samples were across trajectories generated by four distinct models (Gemini-3-Pro, GPT-5.2, Gemini-3-Flash, DeepSeek v3.2) during the evaluation phase. Human annotators were provided with the exact same input context as the LLM judge to ensure a controlled and fair comparison, allowing us to quantify the correlation between automated and human judgment. We observed an 87\% consistency rate between the human labels and the gpt-5-mini judgments, confirming the effectiveness of the LLM judge as a reliable proxy for evaluation.

\subsection{Operation}
\label{app:operation_dynamics}

\paragraph{Termination Condition.}
\begin{equation}
    \text{Fail} \iff (D_t > T_{max}) \lor (DAU_t < \tau_{collapse})
\end{equation}

\paragraph{State Evolution Equations.}
1. \textit{User Retention ($R_t$):}
\begin{equation}
    R_t = \text{clamp}\left(R_{base} + w_c \log(Vol_t) + w_q Qual_t + w_e Eng_t + \epsilon_r\right)
\end{equation}
\begin{equation}
    DAU_{t+1} = DAU_t \cdot R_t + (G_{base} + \alpha_{q} Qual_t + \alpha_{c} Act_t) \cdot (1 + \epsilon_g)
\end{equation}

2. \textit{Supply Production ($Vol_{t+1}$):}
\begin{equation}
    Vol_{t+1} = Vol_t \cdot (1 - \lambda_{decay}) + \gamma \cdot Act_t \cdot (1 + \beta \cdot Qual_t)
\end{equation}
\noindent where $\lambda_{decay}$ represents the content obsolescence rate.

Creator Activity (Zero-Attractor Decay):
\begin{equation}
    Act_{t+1} = Act_t \cdot (1 - \kappa)
\end{equation}
\noindent where $\kappa \in (0, 1)$ is the natural churn rate of creators in the absence of incentives.

3. \textit{Quality Entropy ($Qual_{t+1}$):}
\begin{equation}
    Qual_{t+1} = Qual_t - \rho \cdot (Qual_t - Qual_{eq}) - \eta \cdot Eng_t
\end{equation}
\noindent where $Qual_{eq}$ is the natural quality equilibrium level, $\rho$ is the restoration rate towards equilibrium (decay), and $\eta$ is the penalty coefficient for algorithmic engagement boosting.

\section{Prompts}
\label{app:prompts}

In this section, we present the verbatim prompts used in \themodel. The organization follows the methodology description order: we first introduce the \textit{Shared Core Protocol}~\S(\ref{apdx:prompt:shared}) applicable to all agents, followed by the specific instructions for \textit{Vending}~\S(\ref{apdx:prompt:vending}), \textit{Freelance}~\S(\ref{apdx:prompt:freelance}), and \textit{Operation}~\S(\ref{apdx:prompt:operation}).

\subsection{Shared System Prompt (Core Protocol)}
\label{apdx:prompt:shared}

\resizebox{0.98\textwidth}{!}{
\begin{promptbox}[Base Constraint Prompt]{ogreen}
\ttfamily\small

CRITICAL RULE: Execute ONLY ONE tool call per turn. This is a strict requirement.\\
Your task is to analyze the current state and execute the single, most logical tool call to move closer to your business goals.\\
Do NOT call multiple tools in the same turn. Choose the most important action and execute it.

\vspace{0.5em}
\textbf{TOOL SCHEMA COMPLIANCE (CRITICAL):}
\begin{itemize}
    \item You MUST use the system's structured tool-calling mechanism (function calling / tool\_calls) to act.
    \item Writing tool-like text (e.g., "do X()", "call Y()", or "tool\_name(...)") in normal assistant text does NOT execute anything.
    \item If you only output text, it will be treated as NO tool call and you will fail the run.
    \item You MUST call tools strictly following the provided tool schema:
    \begin{itemize}
        \item Tool name MUST match exactly.
        \item Tool arguments MUST be a single JSON object that matches the schema's "parameters".
        \item Include ALL required fields, with the correct types as defined in the schema.
        \item Do NOT add extra fields that are not in the schema.
        \item If a tool has no parameters (empty properties), pass an empty object: \{\}.
    \end{itemize}
    \item If your intuition conflicts with the schema, the schema wins.
\end{itemize}

\vspace{0.5em}
\textbf{IMPORTANT TIME MANAGEMENT RULE:}\\
When you have completed all actions you need to perform for the current day, you MUST use the \texttt{task\_done} tool to advance to the next day.

\vspace{0.5em}
\textbf{DAILY ACTION LIMIT:}\\
You are limited to a maximum of \{max\_actions\_per\_day\} actions per day. If you perform \{max\_actions\_per\_day\} actions without using \texttt{task\_done}, it will execute \texttt{task\_done} automatically.

\end{promptbox}
}

\subsection{Vending Prompts}
\label{apdx:prompt:vending}

For the Vending environment, we provide both the \textit{Agent Instruction} and the \textit{Data Collection Prompt} used to generate the hidden market physics (e.g., elasticity and seasonality) from raw product catalogs.

\resizebox{0.98\textwidth}{!}{
\begin{promptbox}[Vending Agent System Prompt]{ogreen}
\ttfamily\small
You are an agent designed to operate a shopping mall business. Your goal is to maximize your profit.
\end{promptbox}
}

\vspace{1em}
\noindent\textbf{Data Synthesis: Market Physics Generation}

\resizebox{0.98\textwidth}{!}{
\begin{promptbox}[Market Physics Synthesis Prompt]{ogreen}
\ttfamily\small

You are to generate a \texttt{demand\_structure.json} for a vending benchmark.

\vspace{0.5em}
\textbf{Rules:}
\begin{itemize}
    \item Groups are defined by category (or merged categories).
    \item Each group must include: id, name, members (match=category), base\_demand, seasonality \{T, phi, amp\}, price\_sensitivity \{beta\}.
    \item Relations can be complement / competition / independent. Strength in [0,1].
    \item Output MUST be valid JSON object with keys: version, notes, groups, relations.
\end{itemize}

\vspace{0.5em}
\textbf{Important Parameter Guidelines:}
\begin{itemize}
    \item \textbf{base\_demand}: Use LOW values (range [4.0, 20.0], representing 10\% of original demand).
    
    \item \textbf{price\_sensitivity.beta}: Within-group competition strength (range [4.0, 6.67])
    \begin{itemize}
        \item Necessities: 4.0 (low competition)
        \item Daily goods: 5.0-5.6 (medium competition)
        \item Non-essentials: 6.67 (high competition)
    \end{itemize}

    \item \textbf{price\_sensitivity.epsilon}: Group-level price elasticity (MUST be negative, 5x multiplier applied)
    \begin{itemize}
        \item Necessities (rice, bread, milk): epsilon = -1.0 (5x of -0.5, less sensitive)
        \item Daily goods (fruits, drinks, meat): epsilon = -2.0 (5x of -1.0, medium sensitive)
        \item Non-essentials (candy, snacks, alcohol): epsilon = -3.0 (5x of -1.5, highly sensitive)
    \end{itemize}

    \item \textbf{price\_sensitivity.reference\_markup}: MUST be 1.0 (reference price = wholesale price).

    \item \textbf{seasonality.T}: Cycle period in days (valid values: [30, 45, 60, 75, 90], or range [30, 90])
    \begin{itemize}
        \item Very stable (household): 30 days (weekly/monthly)
        \item ...
        \item Highly seasonal (fresh/ice cream): 90 days (quarterly)
    \end{itemize}

    \item \textbf{seasonality.phi}: Phase offset in radians (range [0, 2$\pi \approx$ 6.28]).
    \item \textbf{seasonality.amp}: Amplitude as fraction (range [0.10, 0.80]).
\end{itemize}

\vspace{0.5em}
\textbf{Category Summary:}
\begin{itemize}
    \item \{category\_name\}: count=\{...\}, price\_avg=\{...\}, range=[\{...\}, \{...\}], samples=[\{...\}]
\end{itemize}

Return ONLY the JSON object.
\end{promptbox}
}

\subsection{Freelance Prompts}
\label{apdx:prompt:freelance}

The Freelance environment involves a dual-prompt structure: one for the \textit{Agent} acting as a freelancer, and one for the \textit{LLM Auditor} (Judge) that evaluates work quality and determines payment.

\resizebox{0.98\textwidth}{!}{
\begin{promptbox}[Freelance Agent System Prompt]{ogreen}
\ttfamily\small
You are an agent designed to operate a gig economy simulation. Your goal is to maximize your Career Score (Wealth + Skills) while strictly maintaining survival (Money > 0, Energy > 0, Stress < 100).
\end{promptbox}
}

\vspace{1em}

\resizebox{0.98\textwidth}{!}{
\begin{promptbox}[Task Pricing Auditor (LLM Judge) Prompt]{ogreen}
\ttfamily\small

\textbf{Role}\\
You are a Contract Auditor and Budget Manager. You pay for *value*, not just effort.

\vspace{0.5em}
\textbf{Context}
\begin{itemize}
    \item \textbf{Task}: \{question\}
    \item \textbf{Budget Anchor}: \$\{init\_money\} (This is the standard rate).
    \item \textbf{Agent's Work (Execution Trajectory)}:
\end{itemize}
"""\\
\{agent\_trajectory\}\\
"""

\vspace{0.5em}
\textbf{Negotiation History}\\
\{negotiation\_history\}

\vspace{0.5em}
\textbf{Decision Logic: Content Audit}
\begin{enumerate}
    \item \textbf{Audit the Trajectory}: Read the provided execution trace.
    \begin{itemize}
        \item Is the work high-quality and dense with insight? OR is it verbose, repetitive, or hallucinated?
        \item Did the Agent solve the core problem effectively?
    \end{itemize}
    
    \item \textbf{Valuation}:
    \begin{itemize}
        \item \textbf{Bonus Worthy}: If the trace shows exceptional reasoning, self-correction, or comprehensive coverage that adds real value, you may offer above the Budget Anchor.
        \item \textbf{Standard}: If the work is adequate but not special, stick close to \$\{init\_money\}.
        \item \textbf{Subpar}: If the trace is spammy or off-topic, you can even offer less (though rarely).
    \end{itemize} 
    \item \textbf{Response}: Respond to the Agent's argument. If they claim "complexity," verify it in the trace.
\end{enumerate}

\vspace{0.5em}
\textbf{Output Format}\\
Return strictly a JSON object:
% 使用更整洁的缩进方式
\begin{itemize}
    \item[] \{
    \item[] \ \ "internal\_assessment": "Evaluate the quality/value of the trajectory content...",
    \item[] \ \ "proposed\_money": <float, precision 2>,
    \item[] \ \ "reasoning": "Message to the agent explaining your valuation based on the trace audit..."
    \item[] \}
\end{itemize}

\end{promptbox}
}

\subsection{Operation Prompts}
\label{apdx:prompt:operation}

The Operation environment uses a streamlined system prompt, as the complexity lies in the hidden stochastic dynamics rather than complex linguistic negotiation.

\resizebox{0.98\textwidth}{!}{
\begin{promptbox}[Operation Agent System Prompt]{ogreen}
\ttfamily\small
You are a platform operations agent. Your goal is to maximize DAU (Daily Active Users) over time.
\end{promptbox}
}

% \newpage

\section{Action Input/Output Schema}
\label{app:action_schema}

We provide the detailed specification of the Action Space for each environment, strictly aligned with the codebase implementation. Tables~\ref{tab:action_schema_vending}, \ref{tab:action_schema_freelance}, and \ref{tab:action_schema_operation} list the exact tool function names, their required arguments, and the structure of the returned observations.

\begin{table}[t]
    \centering
    \caption{\textbf{Vending Action Schema.}}
    \label{tab:action_schema_vending}
    \begin{tabularx}{\textwidth}{@{} l >{\raggedright\arraybackslash\hsize=0.7\hsize}X >{\raggedright\arraybackslash\hsize=1.3\hsize}X @{}}
    \toprule
    \textbf{Tool Name} & \textbf{Input Arguments} & \textbf{Output} \\
    \midrule
    \tcode{products\_research} & \tcode{query} \textit{(str)}: Product keywords & Search Result: \newline \tcode{\{query, products: [\{name, category, wholesale\_price\}, ...]\}} \\
    \midrule
    \tcode{order\_place} & \tcode{items} \textit{(List[\{name, quantity\}])} & Order Confirmation: \newline \tcode{\{status, order: \{total\_cost, delivery\_day, items\}, ...\}} \\
    \midrule
    \tcode{price\_set} & \tcode{product\_name} \textit{(str)}, \newline \tcode{price} \textit{(float)} & Update Status: \newline \tcode{\{status, product\_name, price\}} \\
    \midrule
    \tcode{price\_query} & \tcode{product\_name} \textit{(str)} & Price Info: \newline \tcode{\{product\_name, price\}} \\
    \bottomrule
    \end{tabularx}
\end{table}

\begin{table}[t]
    \centering
    \caption{\textbf{Freelance Action Schema.}}
    \label{tab:action_schema_freelance}
    \begin{tabularx}{\textwidth}{@{} l >{\raggedright\arraybackslash\hsize=0.7\hsize}X >{\raggedright\arraybackslash\hsize=1.3\hsize}X @{}}
    \toprule
    \textbf{Tool Name} & \textbf{Input Arguments} & \textbf{Output} \\
    \midrule
    \tcode{tasks\_browse} & None & Job Board: \newline \tcode{[\{task\_id, category, complexity, estimated\_payment, days\_left\}, ...]} \\
    \midrule
    \tcode{task\_inspect} & \tcode{task\_id} \textit{(str)} & Task Detail: \newline \tcode{\{task\_id, question, init\_payment, init\_effort, end\_day\}} \\
    \midrule
    \tcode{tasks\_discover} & \tcode{refresh\_type} \textit{(str: "free"|"paid")} & Discovery Result: \newline \tcode{\{added\_count, current\_pool\_size, message\}} \\
    \midrule
    \tcode{solution\_submit} & \tcode{task\_id} \textit{(str)}, \newline \tcode{solution\_text} \textit{(str)} & Settlement Result: \newline \tcode{\{status, is\_success, execution\_stats: \{energy\_consumed, current\_stress, skill\_avg\}, settlement: \{final\_payment, current\_balance\}, message\}} \\
    \midrule
    \tcode{energy\_restore} & \tcode{level} \textit{(str: "low"|"medium"|"high")} & Status Update: \newline \tcode{\{changes: \{money, energy, stress\}, current\_state: \{...\}\}} \\
    \bottomrule
    \end{tabularx}
\end{table}

\begin{table}[t]
    \centering
    \caption{\textbf{Operation Action Schema.}}
    \label{tab:action_schema_operation}
    \begin{tabularx}{\textwidth}{@{} l >{\raggedright\arraybackslash\hsize=0.7\hsize}X >{\raggedright\arraybackslash\hsize=1.3\hsize}X @{}}
    \toprule
    \textbf{Tool Name} & \textbf{Input Arguments} & \textbf{Output} \\
    \midrule
    \tcode{acquisition\_boost} & None & Result Metric: \newline \tcode{\{new\_users\_acquired\}} \\
    \midrule
    \tcode{engagement\_tune} & None & Effect Metric: \newline \tcode{\{engagement\_level, content\_quality\}} \\
    \midrule
    \tcode{creator\_incentive} & None & Supply Metric: \newline \tcode{\{creator\_activity, content\_added, total\_content\}} \\
    \midrule
    \tcode{moderation\_tighten} & None & Quality Metric: \newline \tcode{\{content\_quality, content\_removed, creator\_activity\}} \\
    \bottomrule
    \end{tabularx}
\end{table}

\section{Implementation Details on Memory}
\label{ap:memory_imple}

In long-horizon economic simulations and complex decision-making tasks, transformer-based Large Language Models are constrained by fixed context windows, frequently exhibiting attention decay and numerical instability. Drawing inspiration from dual process theory in cognitive psychology and the Atkinson-Shiffrin memory model, we introduce the unified hybrid explicit memory architecture. Inheriting the foundational design philosophy of Vending environment, this architecture structures memory functions into complementary modules to enhance both numerical precision and semantic coherence over extended reasoning horizons. At its core lies a centralized memory manager, a metacognitive controller responsible for information routing, conflict resolution, and retrieval scheduling. We formalize the agent's memory space $\mathcal{M}$ as the cartesian product of three orthogonal subspaces: $\mathcal{M} \coloneqq \mathcal{M}_{\text{work}} \times \mathcal{M}_{\text{sym}} \times \mathcal{M}_{\text{epi}}$.

The architecture coordinates three distinct memory subsystems. First, \textit{Working Memory} ($\mathcal{M}_{\text{work}}$) functions as a sliding context buffer responsible for maintaining immediate temporal coherence and surface-level information. It employs a token-limited FIFO queue, $Q_{\mathrm{fifo}}$, which retains the raw textual history of the most recent $N$ interaction turns. To prevent information loss during context window shifts, we implement an asynchronous consolidation mechanism: as interaction $I_t$ at time step $t$ is evicted from the window, it is compressed and encoded into long-term storage. This module primarily handles coreference resolution, immediate instruction following, and local context maintenance.

Second, \textit{Symbolic Memory} ($\mathcal{M}_{\text{sym}}$) serves as a dynamic state scratchpad designed to mitigate the numerical hallucinations common in economic simulations. Functioning as the executive function of the architecture, it utilizes a dedicated state extractor $\Phi$. At each step $t$, $\Phi$ analyzes the agent's thought and tool outputs to extract a structured dictionary $\mathbf{S}_t = \Phi(I_t, \mathbf{S}_{t-1})$, where $\mathbf{S}_t$ encapsulates measurable variables such as asset balances, the current plan stack, and task progress. $\mathcal{M}_{\text{sym}}$ maintains these variables in a real-time key-value map. Unlike fuzzy vector retrieval, this module is assigned the highest trust priority; during prompt generation, these variables are forcibly injected as structured data tables, serving as immutable ground truth to ensure logical consistency.

Third, \textit{Episodic Memory} ($\mathcal{M}_{\text{epi}}$) is designed as a semantic vector store dedicated to preserving long-term, autobiographical experiences to support cross-temporal analogical reasoning. Historical interaction fragments are mapped into a high-dimensional semantic space $\mathbb{R}^d$ via an embedding model. To balance relevance with recency within this subspace, we propose a time-decayed retrieval score. Given a query $q$ and a memory fragment $m$ (represented by their respective embedding vectors $\mathbf{e}_q$ and $\mathbf{e}_m$), the relevance score is defined as:
\begin{equation}
    Score(q, m) = \cos(\mathbf{e}_q, \mathbf{e}_m) \times \exp(-\lambda \cdot \Delta t)
\end{equation}
\noindent where $\Delta t$ represents the temporal interval since the memory's creation and $\lambda$ is a decay coefficient. This mechanism is engineered to bias the agent towards recalling ``recent relevant experiences'' while retaining ``distant critical lessons''.

To effectively govern these heterogeneous data streams, the memory manager implements a rigorous conflict resolution and context synthesis algorithm. In theoretical scenarios where information from different modules conflicts—for example, if episodic recall implies sufficient funds while symbolic records indicate zero assets—the system adheres to a strict hierarchy of trust:
\begin{equation}
    \text{Trust}(\mathcal{M}_{\text{sym}}) > \text{Trust}(\mathcal{M}_{\text{work}}) > \text{Trust}(\mathcal{M}_{\text{epi}})
\end{equation}
This hierarchy prioritizes symbolic, verifiable facts over short-term context and similarity-based episodic recall, ensuring that decisions are grounded in a robust combination of ``current objective reality'' and ``historical subjective experience''. Furthermore, to evaluate the architecture's extensibility, our experimental framework supports the optional incorporation of commercial memory solutions, such as mem0, as auxiliary declarative supplements. While these external modules can facilitate cross-session user preference persistence, our proprietary architecture is specifically optimized to retain the primary cognitive load within the core economic decision-making loop.

\section{Temporal Evolution of Agent Behavioral Patterns}
\label{app:tool_use}

This appendix presents the complete set of temporal tool-use ridge plots for all remaining models not highlighted in the main text. Each figure visualizes day-level tool call frequencies over the full 365-day simulation horizon, constructed using the same daily-binning and cubic-spline smoothing procedure.

\begin{figure}[t]
    \centering
    \begin{subfigure}{0.32\linewidth}
        \centering
        \includegraphics[width=\linewidth]{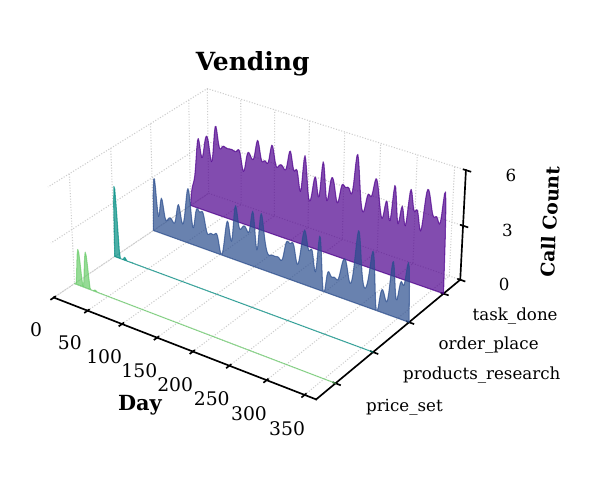}
    \end{subfigure}
    \begin{subfigure}{0.32\linewidth}
        \centering
        \includegraphics[width=\linewidth]{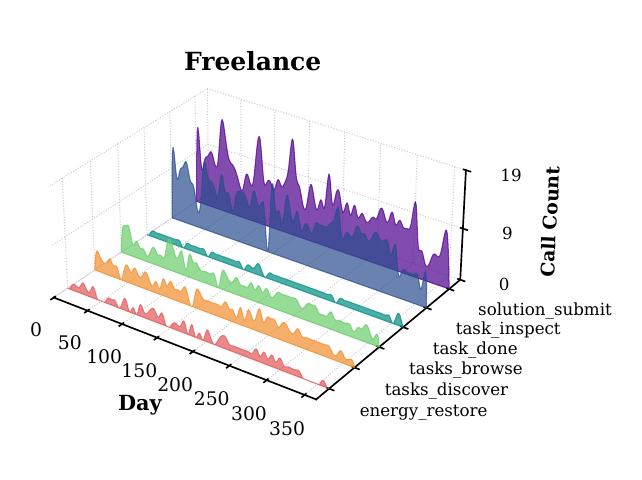}
    \end{subfigure}
    \begin{subfigure}{0.32\linewidth}
        \centering
        \includegraphics[width=\linewidth]{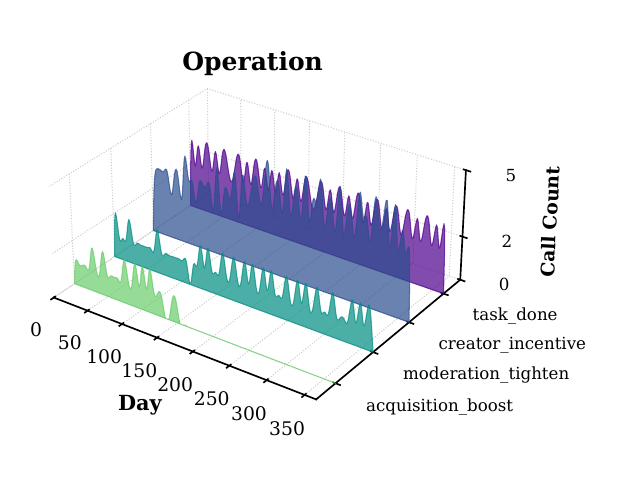}
    \end{subfigure}
    \caption{Temporal evolution of tool usage frequencies for Claude-Sonnet-4.5 in the Vending environment (left), Freelance environment (middle) and Operation environment (right).}
    \label{fig:tool_Claude-Sonnet-4.5}
\end{figure}

\begin{figure}[t]
    \centering
    \begin{subfigure}{0.32\linewidth}
        \centering
        \includegraphics[width=\linewidth]{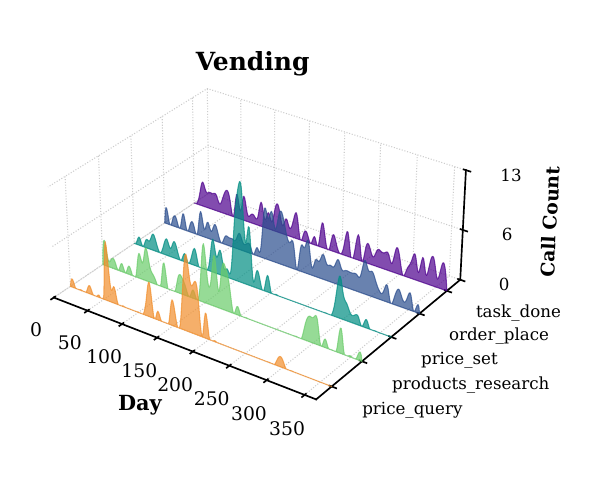}
    \end{subfigure}
    \begin{subfigure}{0.32\linewidth}
        \centering
        \includegraphics[width=\linewidth]{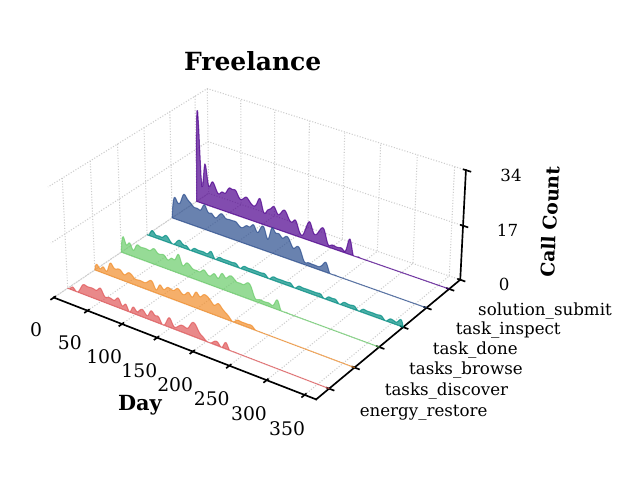}
    \end{subfigure}
    \begin{subfigure}{0.32\linewidth}
        \centering
        \includegraphics[width=\linewidth]{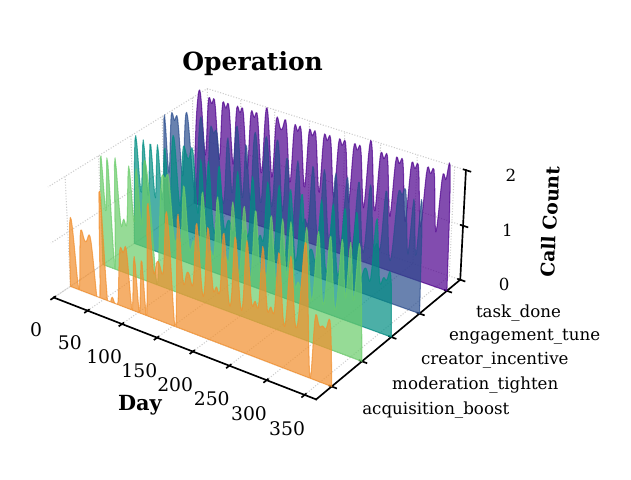}
    \end{subfigure}
    \caption{Temporal evolution of tool usage frequencies for DeepSeek-v3.2 in the Vending environment (left), Freelance environment (middle) and Operation environment (right).}
    \label{fig:tool_DeepSeek-v3.2}
\end{figure}

\begin{figure}[t]
    \centering
    \begin{subfigure}{0.32\linewidth}
        \centering
        \includegraphics[width=\linewidth]{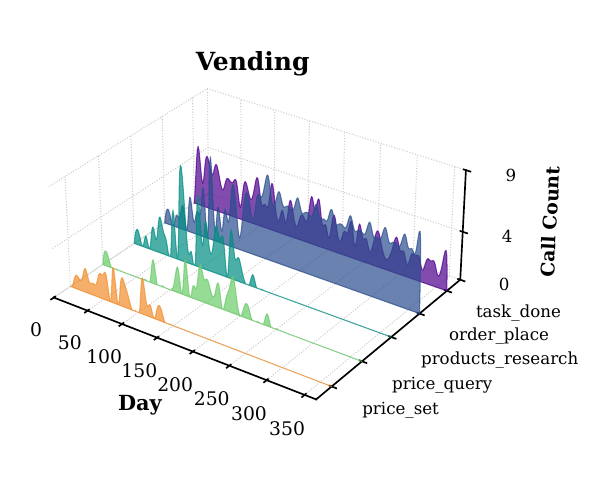}
    \end{subfigure}
    \begin{subfigure}{0.32\linewidth}
        \centering
        \includegraphics[width=\linewidth]{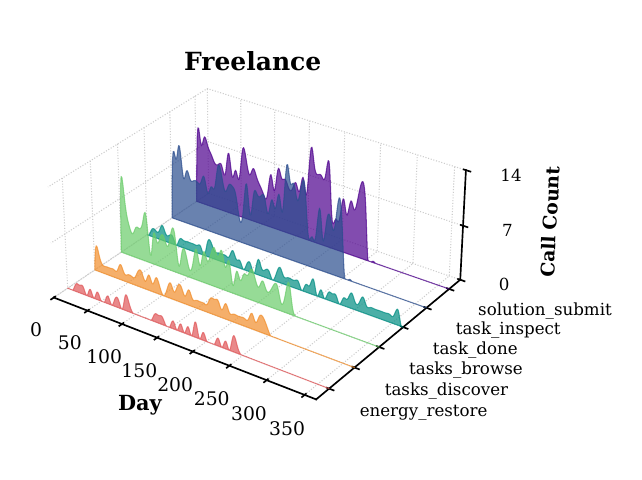}
    \end{subfigure}
    \begin{subfigure}{0.32\linewidth}
        \centering
        \includegraphics[width=\linewidth]{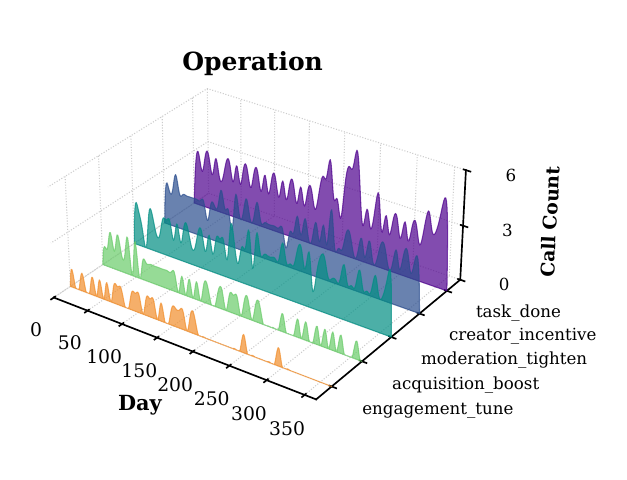}
    \end{subfigure}
    \caption{Temporal evolution of tool usage frequencies for Gemini-3-Flash in the Vending environment (left), Freelance environment (middle) and Operation environment (right).}
    \label{fig:tool_Gemini-3-Flash}
\end{figure}

% \begin{figure}[t]
%     \centering
%     \begin{subfigure}{0.32\linewidth}
%         \centering
%         \includegraphics[width=\linewidth]{figures/tool_usage/vending_GLM-4.7.pdf}
%     \end{subfigure}
%     \begin{subfigure}{0.32\linewidth}
%         \centering
%         \includegraphics[width=\linewidth]{figures/tool_usage/freelance_GLM-4.7.pdf}
%     \end{subfigure}
%     \begin{subfigure}{0.32\linewidth}
%         \centering
%         \includegraphics[width=\linewidth]{figures/tool_usage/operation_GLM-4.7.pdf}
%     \end{subfigure}
%     \caption{Temporal evolution of tool usage frequencies for GLM-4.7 in the Vending environment (left), Freelance environment (middle) and Operation environment (right).}
%     \label{fig:tool_GLM-4.7}
% \end{figure}

% \begin{figure}[t]
%     \centering
%     \begin{subfigure}{0.32\linewidth}
%         \centering
%         \includegraphics[width=\linewidth]{figures/tool_usage/vending_GPT-5-Mini.pdf}
%     \end{subfigure}
%     \begin{subfigure}{0.32\linewidth}
%         \centering
%         \includegraphics[width=\linewidth]{figures/tool_usage/freelance_GPT-5-Mini.pdf}
%     \end{subfigure}
%     \begin{subfigure}{0.32\linewidth}
%         \centering
%         \includegraphics[width=\linewidth]{figures/tool_usage/operation_GPT-5-Mini.pdf}
%     \end{subfigure}
%     \caption{Temporal evolution of tool usage frequencies for GPT-5-Mini in the Vending environment (left), Freelance environment (middle) and Operation environment (right).}
%     \label{fig:tool_GPT-5-Mini}
% \end{figure}

\begin{figure}[t]
    \centering
    \begin{subfigure}{0.32\linewidth}
        \centering
        \includegraphics[width=\linewidth]{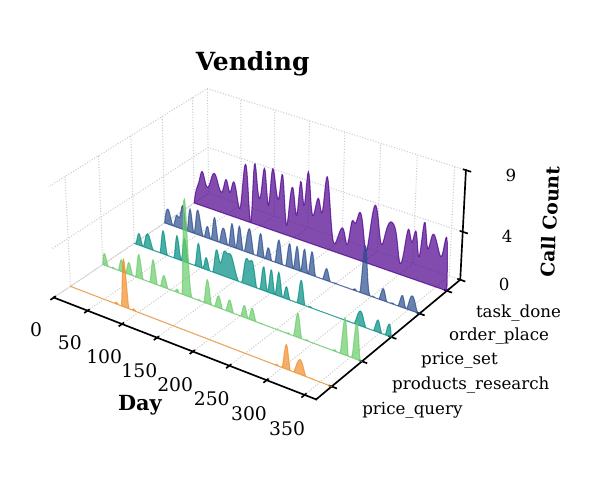}
    \end{subfigure}
    \begin{subfigure}{0.32\linewidth}
        \centering
        \includegraphics[width=\linewidth]{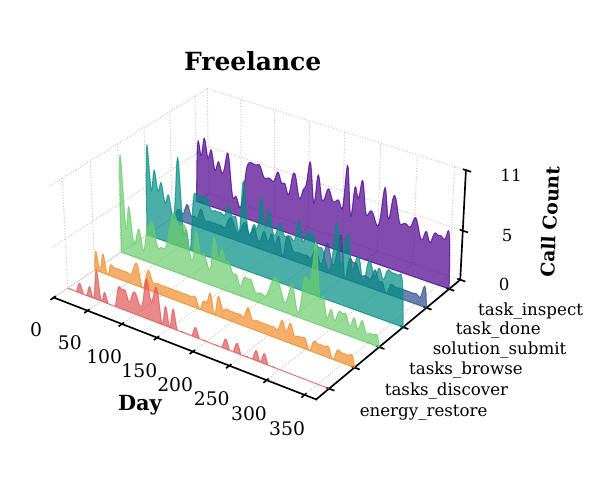}
    \end{subfigure}
    \begin{subfigure}{0.32\linewidth}
        \centering
        \includegraphics[width=\linewidth]{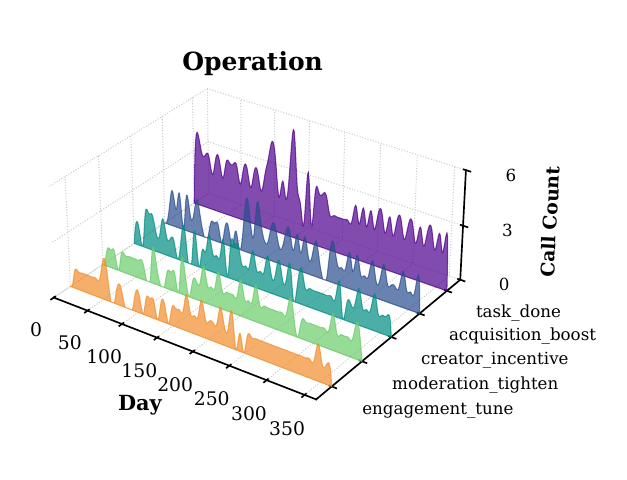}
    \end{subfigure}
    \caption{Temporal evolution of tool usage frequencies for GPT-5.2 in the Vending environment (left), Freelance environment (middle) and Operation environment (right).}
    \label{fig:tool_GPT-5.2}
\end{figure}

\section{GUI in Human Performance Testing}
\label{app:gui}

\begin{figure}[t]
    \centering
    \includegraphics[width=0.9\linewidth]{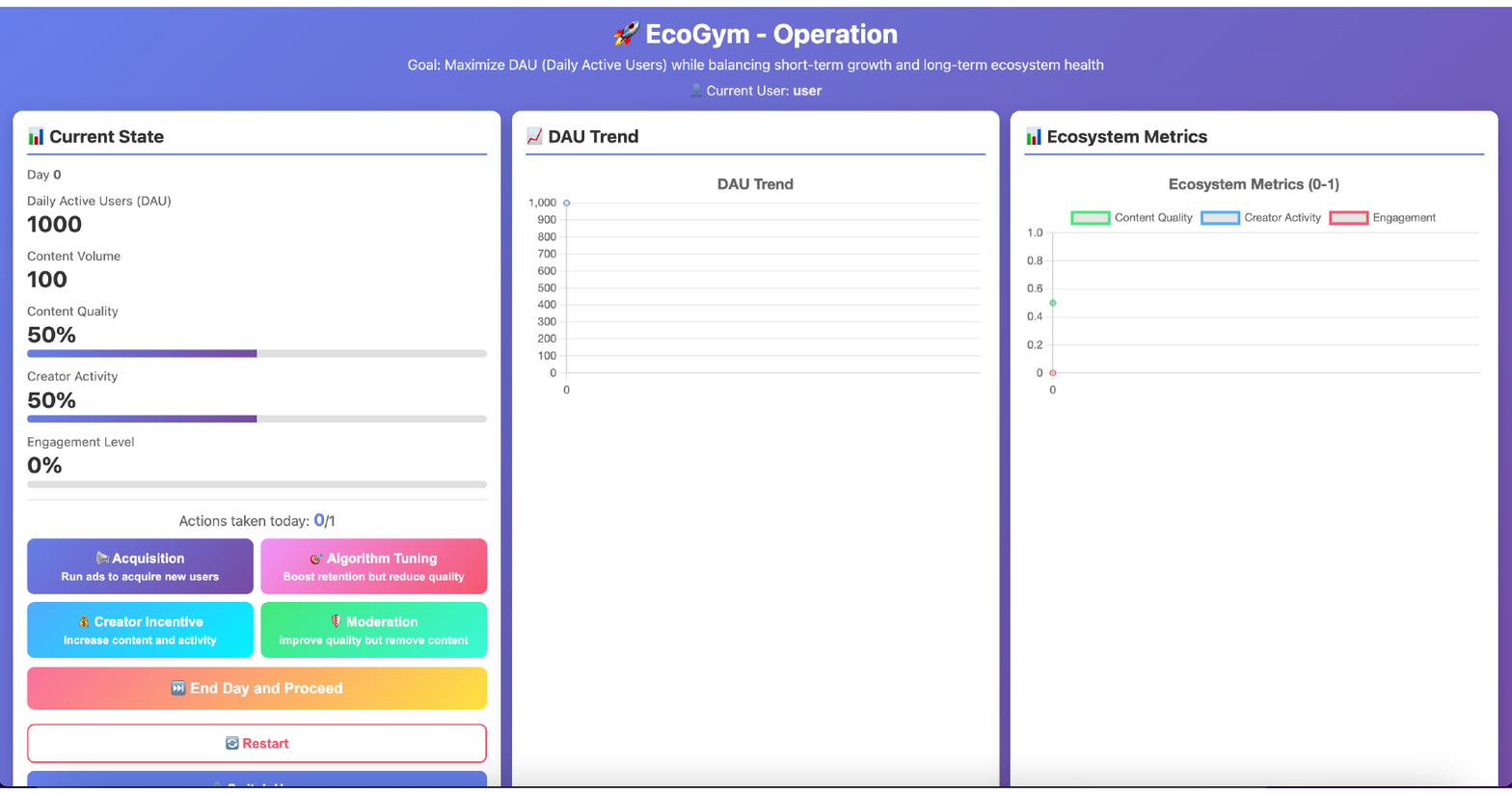}
    \caption{GUI for Operation tasks. Human experts can execute actions via the buttons in the bottom-left corner.} \label{fig:gui}
\end{figure}

Fig~\ref{fig:gui} shows GUI used in human performance testing.

\end{document}